\PassOptionsToPackage{table}{xcolor}
\documentclass[runningheads]{llncs}

\usepackage{eccv}

\usepackage{eccvabbrv}

\usepackage{graphicx}
\usepackage{booktabs}
\usepackage{overpic}
\usepackage[accsupp]{axessibility}  
\usepackage{colortbl}

\newcommand{\methodname}{EyeControl\xspace}
\newcommand{\benchname}{ControlArt-Bench\xspace}
\definecolor{mygray}{gray}{.9}
\definecolor{codegray}{rgb}{0.55,0.57,0.55}
\newcommand*\samethanks[1][\value{footnote}]{\footnotemark[#1]}
\newcommand{\q}[1]{\textcolor{BrickRed}{#1}}
\newcommand{\two}[1]{\textcolor{blue!98!black}{#1}}
\definecolor{myhl}{RGB}{255,230,180}   %
\newcommand{\hlcell}[1]{\cellcolor{myhl}#1}
\newcommand{\hltext}[1]{\setlength{\fboxsep}{1pt}\colorbox{myhl}{#1}}
\definecolor{myhlgreen}{RGB}{220,245,220}
\newcommand{\hlcellgreen}[1]{\cellcolor{myhlgreen}#1}
\newcommand{\hltextgreen}[1]{\setlength{\fboxsep}{1pt}\colorbox{myhlgreen}{#1}}

\usepackage{hyperref}
\usepackage{adjustbox}
\usepackage[symbol]{footmisc}
\usepackage{orcidlink}
\usepackage{wrapfig}
\usepackage{multirow}
\begin{document}

\title{Dotting the Eye: An Intent-Driven Image Retouching Agent for Visual Focus Enhancement} 

\titlerunning{EyeControl}

\author{Chujie Qin\inst{1,2}\thanks{\small{Equal Contribution. This project is done during Chujie Qin's internship at DJI.}}\and
Zilong Zhang\inst{1,2}\samethanks \and
Zewei Chang\inst{1,2} \and
Chunle Guo\inst{1,2} \and
Ruixing Wang\inst{4}\thanks{\small{Corresponding Author.}}\and
Tao Hu\inst{4}\and
Ming-Ming Cheng\inst{1,2,3} \and
Chongyi Li\inst{1,2}\thanks{\small{Project Lead.}}}

\authorrunning{Qin et al.}

\institute{VCIP,CS,Nankai University  \and
NKIARI, Shenzhen Futian \and
AAIS, Nankai University \\
\email{\{chujie.qin,zhangzilong\}@mail.nankai.edu.cn}\\
\email{\{guochunle, cmm, lichongyi,\}@nankai.edu.cn}
\and
DJI Technology Co., Ltd \\
\email{ruixingw@hustunique.com} \\
\email{hubert.hu@dji.com} \\
}
\maketitle


\begin{figure*}[h]
  \centering
  \begin{overpic}[width=\textwidth]{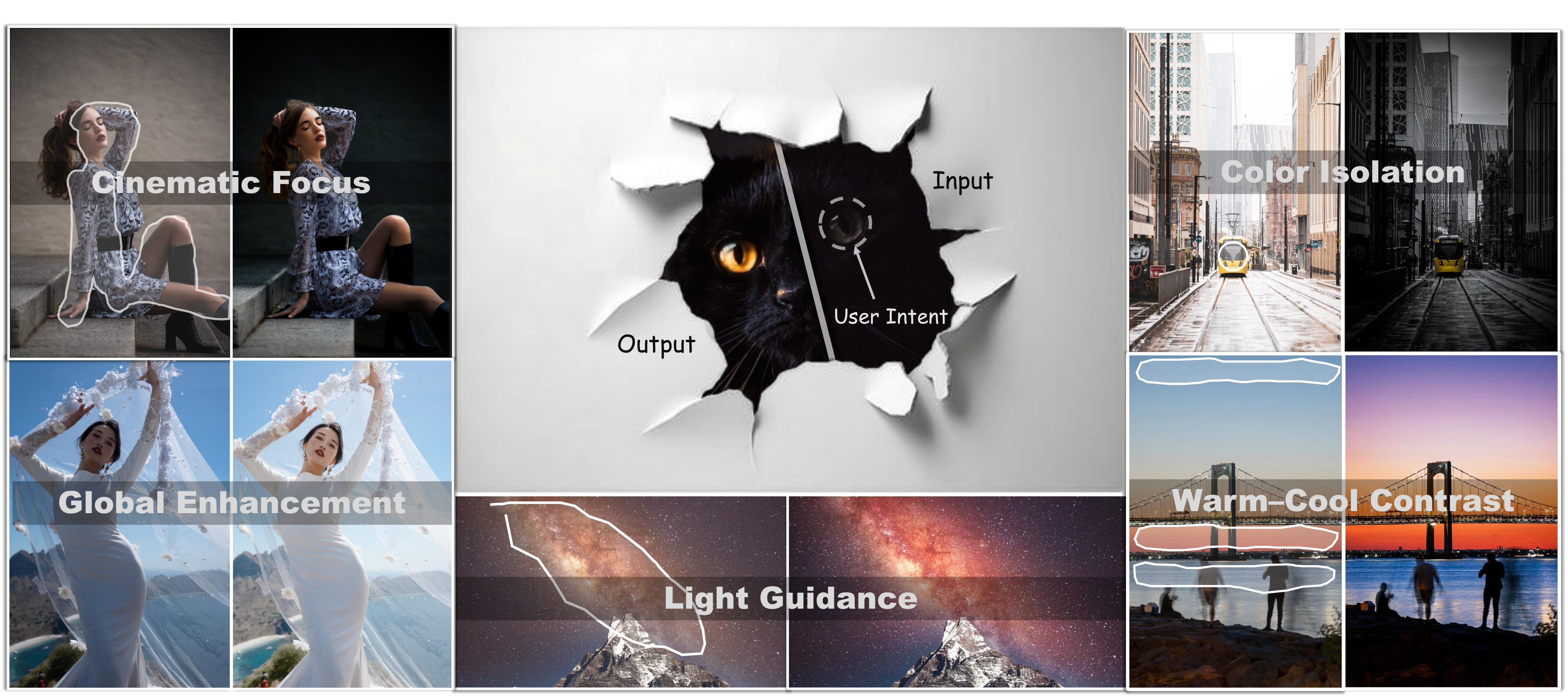}
  \end{overpic}
  \vspace{-7mm}
  \caption{
\textbf{EyeControl} focuses on highlighting the user-intended visual focus through image retouching. With just a few clicks or casual strokes, EyeControl can deliver professional retouching looks—\textit{Cinematic Focus}, \textit{Color Isolation}, \textit{Light Guidance},\textit{Warm–Cool Contrast}, \textit{Global Enhancement}—and, of course, please dot the eye to make the cat "break through" the background wall and pop forward as the new visual focus.
}
  \label{fig:teaser}
\end{figure*}
\renewcommand{\thefootnote}{\arabic{footnote}}

\setcounter{footnote}{0}
\vspace{-10mm}
\begin{abstract}
    Image retouching is commonly formulated as enhancing overall visual quality through color adjustment, but in practice, it also serves to emphasize visual focus by guiding viewers’ attention toward a specific subject or region.
    Achieving such focus-oriented retouching is inherently challenging, as it requires well-coordinated global and local adjustments to manipulate perceptual saliency while maintaining visual naturalness. This intricate process typically demands substantial professional expertise.
    In this study, we propose \textbf{\methodname}, a multi-modal large language model (MLLM)-driven agent with a diffusion-based retouching executor that enables visual focus enhancement under weak user intent. 
    With only a few clicks or coarse strokes, EyeControl directs visual attention to the intended region, effectively “dotting the eye” of the image. 
    The core idea is to explicitly link the weak user intention with the target editing region and the corresponding tonal adjustment operations during retouching. 
    To achieve this, the system first interprets the intent and image content to infer the visual focus and generate structured intent guidance for the retouching executor. 
    Second, the retouching executor is encouraged to respond more strongly to the target region, explicitly aligning its attention map with a designed pseudo-intent map. 
    We also introduce an operation-consistency constraint to improve coordination between global and local adjustments, achieving more natural and coherent retouching. 
    Additionally, we contribute ControlArt-Bench, a high-quality evaluation dataset for visual focus enhancement. 
    Extensive evaluations demonstrate that EyeControl yields perceptually appealing results with stronger intent alignment. Code will be released at \href{https://github.com/DragonisCV/EyeControl}{https://github.com/DragonisCV/EyeControl}.
\end{abstract}

\section{Introduction}
\label{sec:intro}
  \vspace{-2mm}
Image retouching has evolved from manual workflows to learning-based methods\cite{li2019hdrnet,lut2,kim2020pienet,curve2021starenhancer} that enhance perceptual quality through tonal and color adjustments. Recent diffusion- and VLM-based models\cite{duan2025diffretouch,chang2026pertouch,lin2025jarvisevoselfevolvingphotoediting,dutt2025monetgpt} further automate such edits via textual prompts. However, retouching also serves to guide viewers' attention toward a visual focus—a purposeful enhancement that current models fail to explicitly model, even though they support localized edits.

Building upon this observation, we study intent-driven visual focus retouching under weak spatial cues. 
Instead of relying solely on text prompts, we adopt sparse spatial interactions, such as clicks or coarse strokes, as intention signals. Text prompts often lack precise spatial grounding for subtle focus manipulation, while real-world editing workflows typically rely on lightweight spatial interactions to indicate regions of interest. Given an input image and a weak intention signal, the goal is to apply tonal and color adjustments that make the intended visual focus easier to notice, while preserving content and perceptual naturalness.

This problem introduces several inherent challenges.
\textbf{First, spatial intention signals are sparse and ambiguous.}
Spatial interactions such as clicks or coarse strokes provide only partial observations of user intention and do not directly specify the desired editing operations. Inferring the intended visual focus from such sparse signals, therefore, requires reasoning over both user interactions and image context.
\textbf{Second, visual focus lacks explicit control mechanisms in existing retouching models.}
While existing retouching models can apply local adjustments, they do not provide explicit ways to translate user intention into visual focus enhancement.
\textbf{Third, effective visual focus enhancement requires coordinated global and local adjustments.}
Enhancing the prominence of a region often requires coordinated adjustments between the focal area and its surrounding regions, rather than isolated local modifications.

To address the above challenges, we propose \textbf{\methodname}, an intent-driven image retouching agent for visual focus enhancement. 
To enable controllable visual focus enhancement, we introduce Pseudo-Intention Attention Alignment (PAA), which explicitly supervises the model’s internal attention using pseudo-intention labels derived from saliency difference maps. This encourages the model to align its attention with intention-driven saliency shifts, allowing focus modulation to be learned during generation rather than relying on indirect color perturbations. Finally, we introduce an operation consistency constraint to improve coordination between global and local adjustments. Specifically, the guidance for visual focus enhancement is decomposed into global and local operations, and enforces consistency between the result of applying them jointly and the results obtained by applying them sequentially (e.g., global-then-local or local-then-global). This encourages the executor to produce more natural and coherent visual focus enhancement.

In summary, our contributions are threefold:
\begin{itemize}

\item We present a visual-focus-centric perspective on image retouching, enabling visual focus enhancement through retouching and producing more expressive and visually appealing results beyond existing methods, as shown in \cref{fig:teaser}

\item We propose \textbf{\methodname}, an intent-driven image retouching framework that integrates intention reasoning, attention-level focus modulation, and cross-scale coordination, enabling controllable and stable visual focus enhancement while preserving global coherence and perceptual naturalness.

\item  We introduce \textbf{\benchname}, the first image retouching benchmark for visual focus enhancement with paired spatial user-intent annotations and focus-oriented evaluation metrics. Extensive experiments show that our method achieves state-of-the-art performance in visual focus enhancement, while remaining competitive on global and local retouching tasks.
\end{itemize}

\section{Related Work}
\subsection{Global Image Retouching}
With the release of large-scale retouching datasets~\cite{five5k,liang2021ppr10k}, deep learning has become a dominant approach for image retouching. 
Some methods model the retouching process through interpretable adjustment operators, such as tone curves~\cite{curve2021curl,curve2021starenhancer,curve2021zerodce,curve2022cudi}, bilateral grid transformations~\cite{li2019hdrnet,gharbi2017deep,duan2025diffretouch}, or lookup tables (LUTs)~\cite{3dlut,yang2022adaint,lut2}, aiming to mimic traditional editing pipelines. 
Another line of work learns image-to-image mappings directly using convolutional networks~\cite{chen2018deep}, Transformers~\cite{wen2024retouchformer}, or diffusion models~\cite{duan2025diffretouch}.
However, most existing approaches treat retouching as a global enhancement problem. 
Moreover, widely used paired datasets mainly contain globally adjusted targets without explicit modeling of user intention or region-specific emphasis~\cite{five5k,liang2021ppr10k}. 
As a result, the learned supervision often reflects an averaged color preference rather than an intention-driven objective, causing models to converge toward globally consistent but perceptually generic solutions.
In contrast, practical editing is often guided by subjective intention, where adjustments aim to reshape perceptual focus rather than merely improve overall appearance.
\subsection{VLM-Driven Image Retouching}
\vspace*{-2mm}
Recent advances in vision-language models (VLMs) have enabled language-driven image retouching frameworks~\cite{lin2025jarvisart,lin2025jarvisevoselfevolvingphotoediting,chang2026pertouch}. 
By leveraging multimodal understanding, these systems interpret user instructions, analyze image content, and generate editing plans. 
Some approaches further employ large language models to decompose complex requests into sequential operations or parameter adjustments~\cite{dutt2025monetgpt,wu2026retouchiq}.
However, these methods primarily focus on semantic reasoning and instruction following. 
The connection between weak spatial intention signals (e.g., clicks or coarse masks) and controlled perceptual saliency redistribution remains indirect, as spatial cues are typically translated into textual instructions or heuristic editing steps.
In contrast, our approach incorporates intention signals directly into the attention mechanism of a diffusion backbone, enabling structured mask–image interaction and explicit supervision of saliency change.
\vspace*{-4mm}
\subsection{Saliency Retargeting}
\vspace*{-2mm}
Saliency retargeting\cite{retargeting2023realistic,retargeting2011wacv,reducing2022deep} studies how image appearance can be modified to redirect visual attention. Early methods rely on computational saliency models and apply handcrafted adjustments—such as color, contrast, or luminance manipulation—to enhance target regions or suppress distractions\cite{saliencyguided,reducing2022deep}. These works demonstrate that appearance modulation can alter perceptual ordering.
However, existing saliency retargeting techniques are largely rule-based and operate outside modern generative editing frameworks. They typically assume explicit target regions and predefined transformations, and are not designed to handle weak or ambiguous intention signals. Moreover, aesthetic plausibility is rarely considered: aggressive color or contrast manipulation may increase saliency but distort object attributes and reduce perceptual realism.
In contrast, we integrate saliency redistribution directly into a diffusion-based editing backbone. By modeling mask-guided attention interactions and supervising the induced saliency difference, our approach enables retouching for visual focus enhancement.
\vspace*{-4mm}
\section{Method}
\vspace*{-2mm}
We begin by presenting the overall workflow of \methodname(\cref{sec:overview}). We then describe the proposed data generation pipeline, which constructs a high-quality dataset containing diverse samples of weak, spatially specified user intent(\cref{sec:data}). Finally, we introduce the core innovations of \methodname, including VLM-based ambiguous intention understanding and guidance, as well as the training framework for the retouching executor(\cref{sec:eyecontrol}). Together, these components enable intention-driven image retouching that effectively enhances the user-specified visual center under weak or ambiguous inputs.
\begin{figure*}[t]
  \centering
  \begin{overpic}[width=\textwidth]{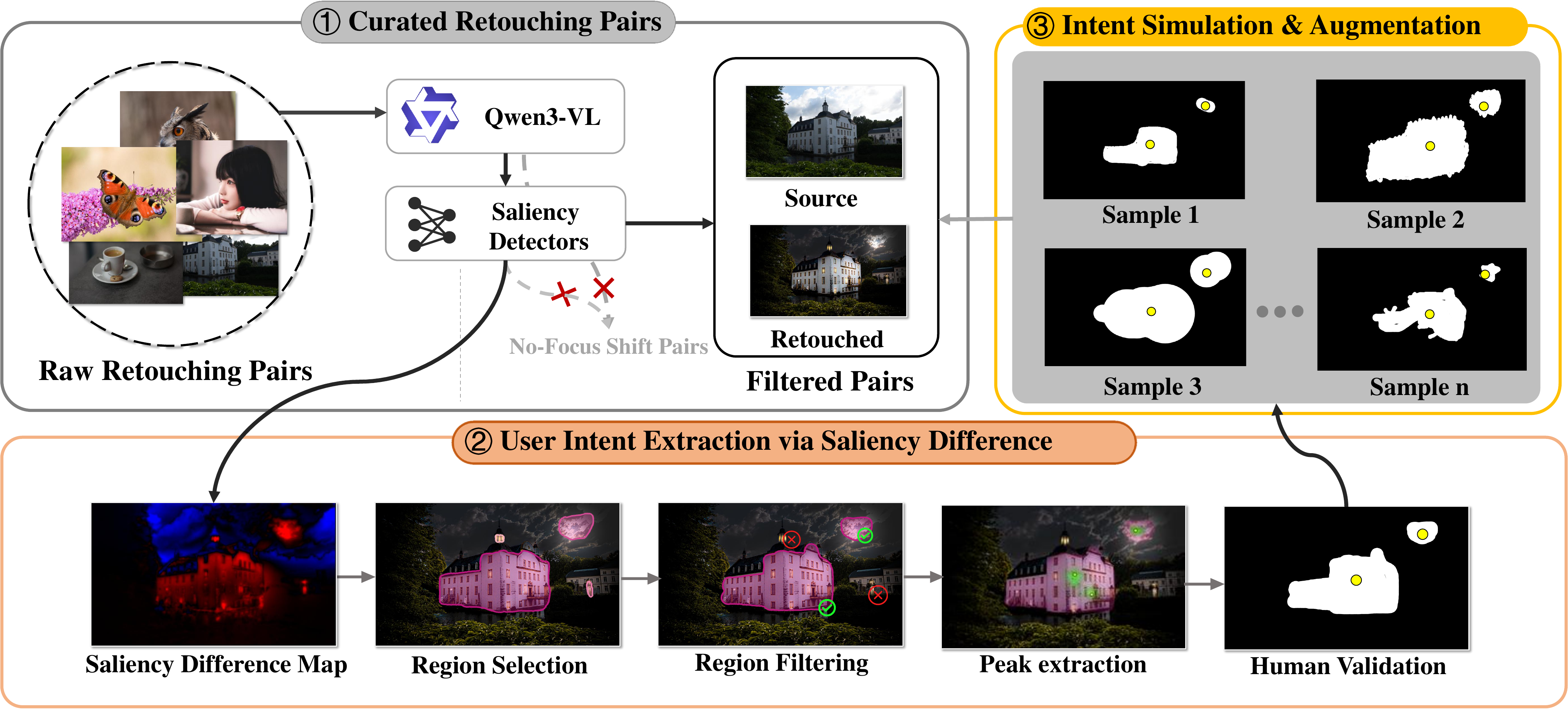}
  \end{overpic}
  \caption{
Pipeline for generating intent-driven retouching supervision:
(1) curating retouching pairs, (2) deriving user intent via saliency difference, and (3) simulating and augmenting intent signals.
  }
  \label{fig:data}
  \vspace*{-6mm}
\end{figure*}
\vspace*{-5mm}
\subsection{ Overview}
\begin{figure*}[t]
  \centering
  \begin{overpic}[width=\textwidth]{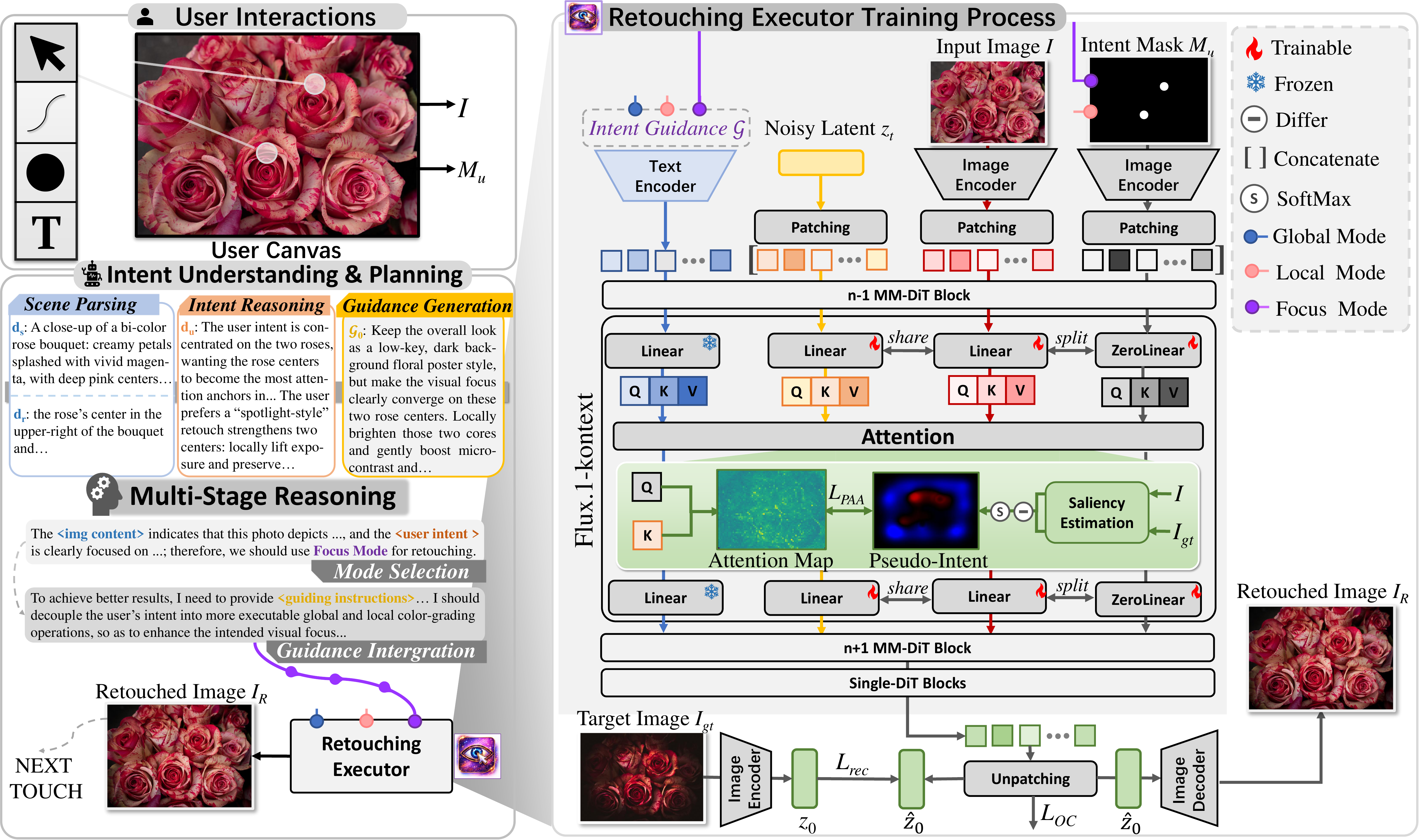}
  \end{overpic}
  \caption{
       The overall workflow of \textbf{\methodname}. Given an image to be retouched, the system first collects user interactions. The \textit{Planner} then performs multi-stage reasoning to parse the image, infer user intent, and select an appropriate execution mode, producing decoupled guidance prompts. Subsequently,  the \textit{Executor} carries out the actual image retouching conditioned on the input image, the inferred intention representation, and the generated guidance. The result is then returned to support iterative refinement.
  }
\vspace*{-6mm}
  \label{fig:main}
\end{figure*}

\vspace*{-2mm}
\label{sec:overview}
\methodname\ is a user-intention-driven image retouching agent that supports not only global enhancement (\emph{Global Mode}) and local adjustment (\emph{Local Mode}), but also visual focus enhancement (\emph{Focus Mode}) based on weak or ambiguous user inputs such as clicks or free-form brush strokes. The overall workflow of \methodname\ is illustrated in \cref{fig:main}. 

The framework consists of three components: a user interaction interface, a VLM-based Planner $\mathcal{A}$, and a Diffusion Transformer-based retouching Executor $\mathcal{E}$. Through the interaction interface, users indicate the desired visual focus using clicks, strokes, or region markings, optionally accompanied by textual guidance $t$ describing their intention. The Planner $\mathcal{A}$ integrates and interprets these ambiguous signals, performs intention reasoning, determines the appropriate execution mode, and generates refined textual guidance for downstream processing. 

Conditioned on the input image $I$, the inferred intention mask $M_u$, and the generated text guidance $\mathcal{G}$, the Executor performs the retouching operation and produces the updated result $I_r$, which can then be fed back into the system for subsequent interaction rounds. Formally, \methodname implements a function: 
\begin{equation}
\mathcal{F}_\theta(I, M_u) \rightarrow I_r.
\end{equation}

\vspace*{-6mm}
\subsection{Data Generation Pipeline}
\label{sec:data}
\vspace*{-2mm}
We design a three-stage data generation pipeline to construct training pairs, with a particular focus on extracting user-intended visual focus, as shown in \cref{fig:data}. Further Details of our train set can be found in the supplemental materials.
 \vspace*{-2mm}
\paragraph{Stage 1: Retouching Pairs Curation.}
Following the practice of JarvisArt\cite{lin2025jarvisart}, we curate a large-scale retouching corpus from PPR10K\cite{liang2021ppr10k}, Lightroom Community, and portfolios of professional retouchers. To ensure that the collected edits indeed induce visual focus enhancement, we apply a two-stage filtering procedure. 
First, Qwen3-VL-32B~\cite{yang2025qwen3technicalreport} is used to remove retouching pairs with negligible perceptual differences or without a distinct visual focus before and after editing. 
Second, we estimate saliency difference between the pre- and post-retouch images using an ensemble of saliency detectors~\cite{saliency1998,saliencytransformer}, and discard samples whose saliency change falls below a predefined threshold. 
This procedure ensures that the retained image pairs present clear visual focus enhancement.
 \vspace*{-3mm}
\paragraph{Stage 2: User-Intent Extraction via Saliency Difference.}
To construct structured supervision for visual focus enhancement, we derive user-intent signals from saliency difference maps computed between reference image pairs. 
Given a saliency difference map $D \in \mathbb{R}^{H \times W}$, we extract polarity-specific responses and 
process them through the following steps. 
1) Region Selection: Determine threshold $\tau$ such that the selected region covers a predefined proportion $\rho$ of total saliency energy, producing an initial binary mask.
2) Region Filtering: Perform connected-component analysis, remove small regions, and retain dominant components based on accumulated energy and size constraints. 
3) Peak Extraction: Extract top-K local maxima within the selected region using non-maximum suppression to obtain sparse intention anchors.
4) Human-in-the-loop validation: select valid intent regions and peaks for training.
\vspace*{-4mm}
\paragraph{Stage 3: Intent Simulation and Augmentation.}
We further introduce an online intent augmentation strategy to better simulate real-world user interaction patterns. Given the intent representation extracted in the previous stage—including coarse regions and salient peaks, we randomly generate diverse interaction masks to simulate typical user inputs, such as point clicks, free-form strokes, and region-based painting. 
While preserving the original intent, this augmentation produces varied interaction patterns and spatial layouts, improving the model’s robustness to different user inputs in retouching.

\subsection{EyeControl Framework}
\label{sec:eyecontrol}
\subsubsection{Intent Understanding and Planning}
\label{sec:agent}
User interactions in our setting are sparse, ambiguous, and often expressed in non-professional language. 
To bridge this gap, we introduce a collaborative multi-VLM agent $\mathcal{A}$, designed as a \textit{Multi-Stage Reasoning} planner that progressively reconstructs the desired visual focus from incomplete observations, as shown in \cref{fig:main}.

Formally, the Planner is modeled as a structured reasoning operator:
\begin{equation}
    \mathcal{G} = \mathcal{A}(I_r, M_u),
\end{equation}
which internally decomposes intent understanding into scene parsing, intent reasoning, and guidance generation before execution planning.
\paragraph{Intent Understanding} The agent first decomposes the scene into complementary semantic channels: a scene-level description $d_s = \phi_s(I_r)$ capturing global stylistic attributes, and a region-aware description $d_r = \phi_r(I_r, M_u)$ focusing on the potential focal area.
The user intent description $d_u$ is inferred as:

\begin{equation}
d_u = \phi(I,d_s, d_r),
\end{equation}
Finally, the planner $\mathcal{A}$ generates an initial unified guidance $\mathcal{G}_0$ based on the scene description $d_s$ and the inferred user intention $d_u$:
\begin{equation}
\mathcal{G}_0 = \psi(d_s, d_u),
\end{equation}
where $\mathcal{\psi}(\cdot)$ denotes the guidance generation function.

\paragraph{Planning} After completing intention understanding, the Planner determines the appropriate adjustment mode based on its joint interpretation of the inferred $\hat{d}_u$ and $d_s$. Specifically, it selects among \emph{Global Mode} for overall tonal and color adjustments, \emph{Local Mode} for region-specific refinement, or \textbf{Focus Mode}, often treated as the default setting, which coordinates both global and local adjustments to enhance the intended visual focus.

To implement this decision, the Planner first produces a unified guidance representation $\mathcal{G}_0$ that captures scene context and inferred intention. This representation is then decoupled into executor-compatible instructions, consisting of a global edit instruction $\mathcal{G}_{\text{global}}$ and a mask-aware local edit instruction $\mathcal{G}_{\text{local}}$. 

The final executor-compatible instruction is formulated as:
\begin{equation}
\mathcal{G} =
m_1 \cdot \{\mathcal{G}_{\text{global}}\}
\cup
m_2 \cdot \{\mathcal{G}_{\text{local}}\},
\end{equation}
where $\mathbf{m}=(m_1,m_2)\in\{(1,0),(0,1),(1,1)\}$ is a binary mode indicator controlling the activation of global and local instructions.
Specifically,

\begin{equation}
\mathbf{m} =
\begin{cases}
(1,0), & \text{Global Mode},\\
(0,1), & \text{Local Mode},\\
(1,1), & \text{Focus Mode}.
\end{cases}
\end{equation}

Under \emph{Focus Mode}, both global and local instructions are activated to collaboratively enhance the intended visual focus. The assembled instruction $\mathcal{G}$ is then fed into the diffusion executor together with the input image $I$ and the intent mask $M_u$. 
In Global Mode, the mask is replaced by an all-zero mask, so that the retouching is driven purely by global guidance.

\subsubsection{Retouching Executor} Next, we describe the architecture and training process of the retouching executor.
\paragraph{Architecture} The overall architecture is built upon Flux.1-Kontext\cite{labs2025flux1kontextflowmatching} and augmented with intention-aware conditioning. Given an input image $I$, a user-provided intent mask $M_u$, and structured intent guidance text $\mathcal{G}$, the image is first encoded into a latent representation via a VAE encoder. During training, the latent is perturbed according to the diffusion process to obtain a noisy latent $z_t$. $I$, $M_u$, $z_t$, and $\mathcal{G}$ are patchified into token sequences and jointly processed by stacked MM-DiT blocks. Within each block, latent tokens interact with image-condition, mask, and text tokens through multi-modal attention. To mitigate feature interference across modalities, the mask branch is equipped with dedicated projection parameters, enabling mask-aware modulation of latent representations. After passing through multiple MM-DiT layers followed by Single-DiT refinement blocks, the network predicts the denoised latent. 
Training is supervised in latent space using a standard Flow-Matching reconstruction loss:

\begin{equation}
L_{rec} = \| \hat{z}_0 - z_0 \|_2^2 ,
\end{equation}
where $\hat{z}_0$ is the predicted clean latent and $z_0$ denotes the encoded latent of the ground-truth retouched image $I_{gt}$. The final retouched output $I_r$ is obtained by decoding the predicted latent through the VAE decoder.
\paragraph{Pseudo-Intent Guided Attention Alignment(PAA)}
\label{sec:loss}
Although mask tokens participate in multi-modal conditioning, standard attention mechanisms do not explicitly regulate how spatial intention influences latent interactions. As a result, mask concatenation alone cannot guarantee that user interaction leads to consistent emphasis on the intended regions during denoising.

In the DiT, attention at timestep $t$ follows the standard formulation:
\begin{equation}
\mathbf{A}_t = \mathrm{Softmax}
\left(
\frac{\mathbf{Q}_t \mathbf{K}_t^\top}{\sqrt{d}}
\right),
\end{equation}
where $\mathbf{Q}_t = W_Q \mathbf{X}_t$ and $\mathbf{K}_t = W_K \mathbf{X}_t$ are the query and key projections of the latent features $\mathbf{X}_t$, and $d$ denote the channel dimension.

When mask conditioning is incorporated, mask tokens derived from $M_u$ are appended to the token sequence. 
Let $\mathbf{Q}_t^{(M)}$ denote queries originating from mask tokens, and $\mathbf{K}_t^{(z)}$ denote keys from noisy latent tokens $z_t$. 
The corresponding mask-to-latent attention map is:
\begin{equation}
\mathbf{A}_t^{(M \rightarrow z)}
=
\mathrm{Softmax}
\left(
\frac{\mathbf{Q}_t^{(M)} \left(\mathbf{K}_t^{(z)}\right)^\top}{\sqrt{d}}
\right).
\end{equation}
This sub-attention determines which spatial regions in the latent representation are influenced by the mask guidance.  
Supervising $\mathbf{A}_t^{(M \rightarrow z)}$ therefore directly constrains the effective region of mask-guided modulation.

We construct a pseudo-intent map $\tilde{A}$ from saliency difference between pre- and post-adjustment signals. 
Unlike a binary mask, $\tilde{A}$ encodes signed prominence variations, indicating both regions to be enhanced and regions to be suppressed.

Let $S(\cdot): \mathbb{R}^{H\times W\times 3} \rightarrow [-1,1]^{H\times W}$ denote a saliency estimation function that produces a normalized saliency map.
The pseudo-intent map is defined as:

\begin{equation}
\tilde{A} = SoftMax(S(I_{gt}) - S(I)),
\end{equation}
where $I$ and $I_{gt}$ denote the input image and the ground-truth retouched image, respectively. 
Unlike a binary mask, $\tilde{A}$ encodes signed prominence variations, indicating both regions to be enhanced and regions to be suppressed.
We align the normalized mask-to-latent attention with the pseudo-intent map:
\begin{equation}
\mathcal{L}_{\text{PAA}}
=
\mathbb{E}_{t}
\left[
\frac{1}{L}
\sum_{\ell=1}^{L}
\left\|
\mathrm{Norm}\big(\mathbf{A}^{(M \rightarrow z),\ell}_t\big)
-
\mathrm{Norm}(\tilde{A})
\right\|_2^2
\right],
\end{equation}
where $t$ denotes the diffusion timestep, $L$ is the number of attention layers, and $\mathbf{A}^{(M \rightarrow z),\ell}_t$ represents the mask-to-latent attention map at layer $\ell$ and timestep $t$. 
$\mathrm{Norm}(\cdot)$ denotes spatial normalization applied to both maps to ensure comparable scales.
Through this alignment, the model learns to associate spatial masks with consistent attention allocation, enabling intent-aware retouching.
\begin{wrapfigure}{l}{0.34\linewidth}
\vspace{-12pt}
\centering
\includegraphics[width=1\linewidth]{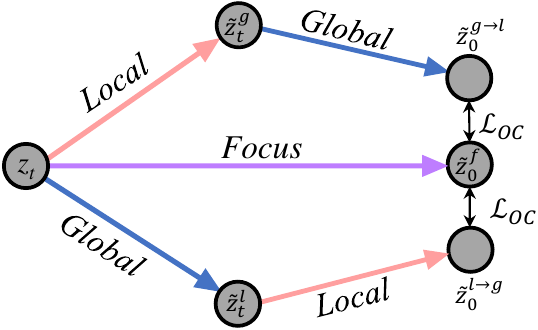}
\caption{Illustration of Operation Consistency Loss.}
\label{fig:oc_loss}
\vspace{-12pt}
\end{wrapfigure}

\vspace{-12pt}
\paragraph{Operation-Consistensy Loss} 
\label{sec:consist} 
Global and local retouching operations form compositional transformations within a unified editing space. 
In particular, samples annotated under the \textit{focus} mode simultaneously contain global and local editing instructions. 
This allows us to decompose a \textit{focus} sample into two independent conditioning signals corresponding to the Global and Local modes.

Ideally, applying these operations sequentially should yield results consistent with applying them jointly under the \textit{focus} condition. 
However, diffusion models trained under mixed modes often exhibit order-sensitive behavior: dominant global adjustments may suppress local guidance, while local edits may disrupt overall visual coherence.

To address this issue, we introduce an \textbf{operation-consistency} objective using \textit{focus} mode samples as supervision. For each \textit{focus} sample, we construct one joint denoising path and two sequential paths with randomly sampled execution orders (\cref{fig:oc_loss}).

For the focus path, the model performs a standard forward pass under the focus mode, producing the latent prediction $\hat{z}^{f}_{t}$ at timestep $t$, which serves as the anchor target.

\emph{Global-to-Local Path.}
We first denoise under the global-only condition to obtain $\hat{z}^{g}_{0}$, then inject noise:
\begin{equation}
\tilde{z}^{g}_{t}=(1-\sigma_t)\hat{z}^{g}_{0}+\sigma_t\epsilon,
\end{equation}
where $\epsilon$ is Gaussian noise and $\sigma_t$ controls the noise injection strength. 
The re-noised latent $\tilde{z}^{g}_{t}$ is refined under the local condition, producing $\hat{z}^{g\rightarrow l}_{0}$.

\emph{Local-to-Global Path.}
Similarly, swapping the execution order yields $\hat{z}^{l\rightarrow g}_{0}$.

Finally, we enforce operation-consistency by aligning the sequential prediction with the Focus anchor:
\begin{equation}
\mathcal{L}_{\text{OC}} =
\lambda \big\|
(\hat{z}^{o}_{0})-\text{stopgrad}(\hat{z}^{f}_{0})
\big\|_2^2,
\end{equation}
where $o\in\{g\!\rightarrow\!l,\;l\!\rightarrow\!g\}$ denotes the execution order sampled during training.

In summary, the overall training objective is:
\begin{equation}
\mathcal{L}=\mathcal{L}_{rec}+\lambda_1\mathcal{L}_{PAA}+\lambda_2\mathcal{L}_{OC}.
\end{equation}

\section{Experiments}
\subsection{ Implementation details}
\subsubsection{Training setting} 
\label{sec:implementation_details}
We utilize Flux-1.0-Kontext\cite{labs2025flux1kontextflowmatching} as the base model. For fine-tuning, we apply LoRA (rank $r = 128$) and optimize the network using a learning rate of $2 \times 10^{-5}$. The training is conducted across 4$\times$H20 for 10,000 steps with a batch size of 1 per GPU.

\vspace{-5mm}
\subsubsection{\benchname}
To the best of our knowledge, there is currently no publicly available retouching benchmark designed to evaluate image editing quality under ambiguous user intentions, particularly for the assessment of visual focus enhancement. Therefore, we introduce \textbf{\benchname}, a high-quality evaluation dataset specifically constructed for focus enhancement. \benchname contains 200 groups of real-user retouching samples, each consisting of an \emph{IntentMask–Image} pair. The dataset spans diverse categories, including portraits, natural landscapes, architecture, food, and animals. To further improve its usability and reproducibility, we additionally provide, for each sample, an editing instruction aligned with the corresponding retouching intention.

\begin{table}[!t]
\vspace{-2mm}
\centering
\renewcommand{\arraystretch}{1.1}
\setlength{\tabcolsep}{4pt}
\scriptsize
\caption{Quantitative evaluation of retouching performance on \benchname. The \textcolor{red}{best} and
\textcolor{blue}{second-best} results are highlighted. FA, PQ, and O denote the metrics evaluated by Qwen2.5-VL-72B~\cite{bai2025qwen25vltechnicalreport}.}
\vspace{-2mm}
\label{table:compare_result}
\begin{tabular}{lcccccccc}
\toprule
Method & PSNR$\uparrow$ & SSIM$\uparrow$ & FA$\uparrow$ & PQ$\uparrow$ & O$\uparrow$ & KL$\downarrow$ & CC$\uparrow$ & SIM$\uparrow$\\
\midrule

\rowcolor{gray!12}
\multicolumn{9}{l}{\textit{\textbf{Advanced Retouching Agents}}}\\
JarvisArt~\cite{lin2025jarvisart} 
& 18.8169 & 0.7463 & 7.8350 & 8.7050 & 8.2087 & \textcolor{blue}{0.2229} & 0.9548 & 0.8785 \\
JarvisEvo~\cite{lin2025jarvisevoselfevolvingphotoediting} 
& 20.9291 & \textcolor{blue}{0.8293} & 7.9700 &9.2000 &8.4895 & 0.2928 & 0.9644 & 0.8913 \\
PerTouch~\cite{chang2026pertouch} 
& 15.9370 & 0.5946 & 7.4650& 8.4100 &7.8240 & 0.2700 & 0.9621 & 0.8828 \\

\midrule
\rowcolor{gray!12}
\multicolumn{9}{l}{\textit{\textbf{Open-Source Editing Diffusion Models}}}\\
UniWorld-v2~\cite{li2025uniworldv2} 
& 18.0759 & 0.6879 & \textcolor{blue}{8.2850}& 8.5850 &8.3832 & 0.7567 & 0.9363 & 0.8494 \\
Step1X-Edit~\cite{stepedit} 
& 18.1433 & 0.7533 & 8.1950& 8.8350 &8.4632 & 0.2508 & 0.9564 & 0.8789 \\
FLUX.1-Kontext~\cite{labs2025flux1kontextflowmatching} 
& 18.0646 & 0.7471 & 8.2200 &7.2150& 7.4000 & 0.8231 & 0.9243 & 0.8362 \\
Qwen-Image-Edit-2511~\cite{wu2025qwenimagetechnicalreport} 
& 18.7092 & 0.7029 & 8.2550& 8.6200 &8.3530 & 0.2610 & 0.9577 & 0.8783 \\

\midrule
\rowcolor{gray!12}
\multicolumn{9}{l}{\textit{\textbf{Commercial Closed-Source Models}}}\\
GPT-Image-1.5~\cite{openai2024gptimage} 
& 16.7133 & 0.5659 & \textcolor{red}{8.3291} &8.8101 &8.5196 & 0.6651 & 0.9141 & 0.8269 \\
Nano-Banana-$2$~\cite{geminiteam2025geminifamilyhighlycapable} 
& \textcolor{blue}{21.1127} & 0.7633 & 8.1800 &\textcolor{blue}{9.2900} & \textcolor{blue}{8.6785} & 0.2389 & \textcolor{blue}{0.9735} & \textcolor{red}{0.9085} \\

\midrule
\textbf{\methodname}(Ours)
& \textcolor{red}{21.8845} 
& \textcolor{red}{0.8517} 
& 8.2100 
& \textcolor{red}{9.3050}
&\textcolor{red}{8.7054}
& \textcolor{red}{0.2215} 
& \textcolor{red}{0.9743} 
& \textcolor{blue}{0.9068} \\

\bottomrule
\end{tabular}
\end{table}

\begin{figure*}[!t]
  \centering
  \begin{overpic}[width=\textwidth]{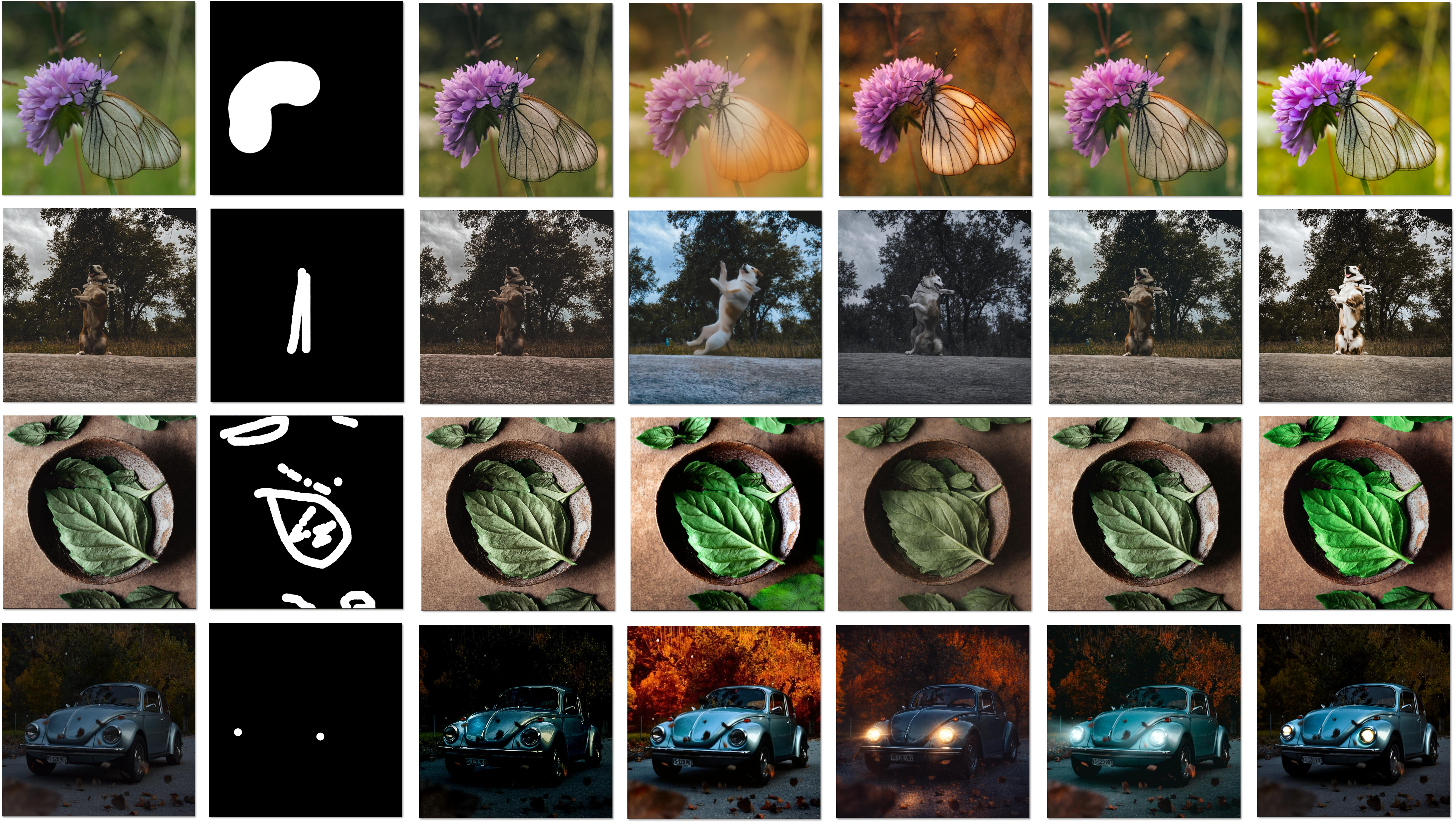}
    \scriptsize
    \put(3.7,-1.5){Input}
    \put(14.8,-1.5){Intent Mask}
    \put(31,-1.5){JarvisEvo}
    \put(43,-1.5){Step1X-Edit}
    \put(56,-1.5){GPT-Image-1.5}
    \put(72,-1.5){Nano-Banana 2}
    \put(90,-1.5){\textbf{Ours}}
  \end{overpic}
  \vspace{-4mm}
  \caption{Qualitative comparisons with other methods on ControlArt-Bench.}
  \vspace{-4mm} 
  \label{fig:compare_result}
\end{figure*}
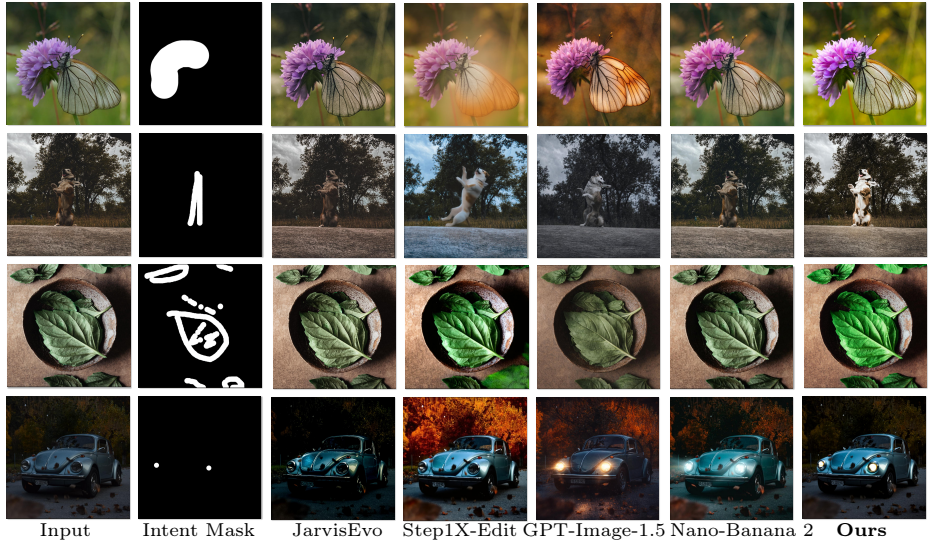

\subsubsection{Metrics}
\vspace{-4mm}
We report eight metrics: PSNR, SSIM, FA, PQ, O, KL, CC, and SIM. PSNR and SSIM measure the overall fidelity and structural similarity to the reference image. We introduce Focus Alignment (FA) to evaluate how well the visual focus in the retouched image aligns with the regions specified by the user intent (0–10 scale). PQ measures contextual coherence and artifacts (0–10 scale)\cite{ku2024imagenhub}. The overall score $\text{O}=\sqrt{\text{FA}\times \text{PQ}}$. Furthermore, KL, CC, and SIM quantify visual-focus (saliency) discrepancies between the retouched and reference images\cite{kummererSaliencyBenchmarkingMade2018}. See the supplementary material for details.
\subsection{ Compared with other Methods }
We compare \methodname\ with open-source agent-based retouching models (JarvisArt\cite{lin2025jarvisart}, JarvisEvo\cite{lin2025jarvisevoselfevolvingphotoediting}, PerTouch\cite{chang2026pertouch}), generative editing models (UniWorld-v2\cite{li2025uniworldv2}, Step1X-Edit\cite{stepedit}\footnote{the latest version of \textit{Step1X-Edit-v1p2}, released on Nov 26, 2025}, FLUX.1-Kontext\cite{labs2025flux1kontextflowmatching}, Qwen-Image-Edit-2511\cite{wu2025qwenimagetechnicalreport}), and commercial systems (GPT-Image-1.5\cite{openai2024gptimage}\footnote{Generative model from OpenAI, released in 
December, 2025}, Nano-Banana 2\cite{geminiteam2025geminifamilyhighlycapable}\footnote{the latest model from Google, released in March, 2026}). To ensure a fair comparison, we provide detailed long-form editing instructions that are carefully aligned with the underlying user intentions. In addition, based on the Intent Mask, we generate corresponding bounding-box location descriptions to accommodate the input formats required by different models. Further implementation details, along with additional qualitative results on global and local image retouching, are provided in the supplementary material.

\begin{figure}[!t]
\vspace{-4mm}
  \centering
  \begin{overpic}[width=\textwidth]{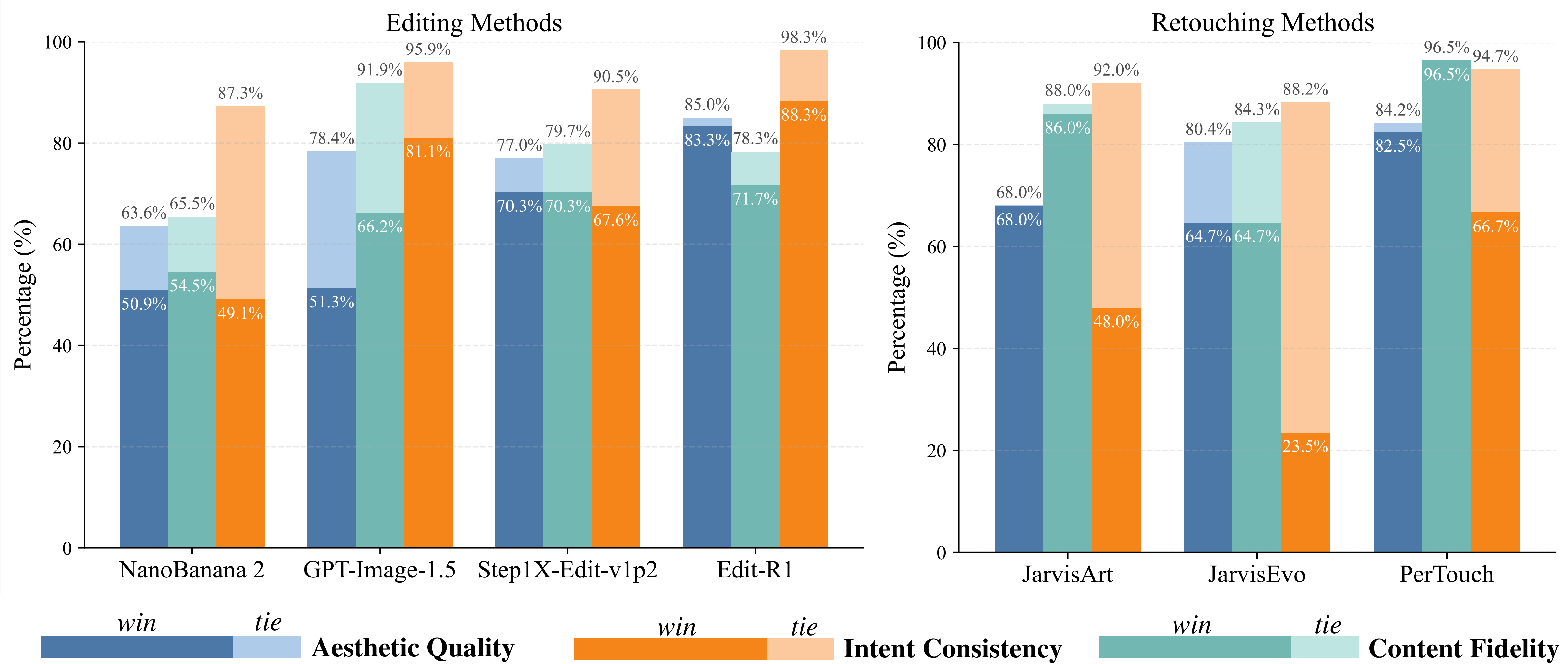}
  \end{overpic}
  \vspace{-3mm}
  \caption{
    \textbf{User Preference Study.}
   Comparison across three criteria: aesthetic quality, intent consistency, and content fidelity.
    We report the percentages of win and tie outcomes of \methodname against each competing method. Results are shown separately for image editing models (left) and agent-based retouching models (right).
  }
  \vspace{-4mm}  
  \label{fig:user_study}
\end{figure}

\paragraph{Comparison on ControlArt-Bench}
As shown in \cref{table:compare_result}, \methodname\ outperforms existing image retouching models, open-source diffusion-based editing models, and GPT-Image-1.5 on the majority of evaluation metrics. Moreover, our approach achieves performance comparable to NanoBanana~2, demonstrating its strong competitiveness against advanced commercial systems. 
We observe in\cref{fig:compare_result} that some editing models score well on the Focus Alignment metric mainly because they introduce large structural changes, rather than focus guidance via subtle retouching. In contrast, our method demonstrates clear advantages in visual focus enhancement. It combines the high content fidelity typically observed in retouching models with the broad editing flexibility characteristic of diffusion-based editing approaches. More importantly, \methodname\ substantially surpasses competing methods in intention alignment, achieving significantly stronger consistency with the specified user intent.

\paragraph{User Preference Study}
Evaluating intent-driven image retouching is subjective, as aesthetic preference varies across individuals. To quantitatively assess this subjectivity, we conduct a large-scale user study on ControlArt-Bench. We recruit 50 participants to compare our method against seven state-of-the-art approaches, including four image editing methods and three image retouching methods.

The evaluation is conducted from three perspectives: 
(1) \textbf{Aesthetic Quality}, measuring whether the image is visually pleasing; 
(2) \textbf{Intention Alignment}, assessing whether the highlighted region aligns with the user’s intended focus; and 
(3) \textbf{Image Fidelity}, evaluating whether the edited result preserves the original content structure.

For each comparison, we randomly select one competing method and perform a blind pairwise comparison against our approach. Participants are asked to choose the better result under each metric (or indicate a tie). As shown in \cref{fig:user_study}, \methodname consistently outperforms all competing approaches across all three evaluation criteria.

\subsection{Ablation Study and Discussion}
\begin{table}[t]
\centering
\tiny
\setlength{\tabcolsep}{1pt}
\renewcommand{\arraystretch}{0.9}
\caption{Quantitative Ablation of EyeControl.}
\begin{tabular}{cc|cccc|cc|cc|cc|cc}
\toprule
\multirow{2}{*}{\textbf{Variants}} &\multirow{2}{*}{\textbf{Methods}}
& \multicolumn{4}{c}{\textbf{Focus}}
& \multicolumn{2}{c}{\textbf{Global}}
& \multicolumn{2}{c}{\textbf{Local}}
& \multicolumn{2}{c}{\textbf{Multi-Round}}
& \multicolumn{2}{c}{\textbf{Average}} \\
\cmidrule(lr){3-6}
\cmidrule(lr){7-8}
\cmidrule(lr){9-10}
\cmidrule(lr){11-12}
\cmidrule(lr){13-14}
&& \textbf{PSNR} & \textbf{SSIM} & \textbf{O} & \textbf{KL}
& \textbf{PSNR} & \textbf{SSIM}
& \textbf{PSNR} & \textbf{SSIM}
& \textbf{PSNR} & \textbf{SSIM}
& \textbf{PSNR} & \textbf{SSIM}\\
\midrule
A & baseline
& 21.62 & 0.8502 & 8.64 & \two{0.2006} 
& \two{18.54}  & 0.7655 
& 27.63 & 0.9428 
& 19.03 & 0.8165
& 21.70 & 0.8438 \\

B& A+Planner 
& 21.71 & \two{0.8565} & 8.64 & 0.2199 
& 18.07 & 0.7596 
& 27.75 & \two{0.9433} 
& 19.48 & 0.8254 
& 21.76 & 0.8462\\ 

C & B+\hltext{PAA}
& \hlcell{\q{21.92}} 
& \hlcell{\q{0.8616}} 
& \hlcell{\two{8.66}} 
& \hlcell{\q{0.1655}} 
& 17.75 
& \two{0.7657} 
& \hlcell{\q{28.55}} 
& \hlcell{\q{0.9569}} 
& \two{20.33} & \two{0.8340} 
& \two{22.14} 
& \two{0.8546}  \\

D & C+\hltextgreen{OCLoss} 
& \two{21.88} 
& 0.8517 
& \q{8.71} 
& 0.2215 
& \hlcellgreen{\q{23.34}} 
& \hlcellgreen{\q{0.800}} 
& \two{28.37} 
& 0.9408  
& \hlcellgreen{\q{20.81}} & \hlcellgreen{\q{0.8353}}
& \hlcellgreen{\q{24.17}} 
& \hlcellgreen{\q{0.8587}}\\
\bottomrule
\end{tabular}
\label{tab:ablation}

\end{table}

\begin{figure}[t]
\vspace{-2mm}
  \centering
  \begin{minipage}[t]{0.38\linewidth}
    \centering
    \includegraphics[width=\linewidth]{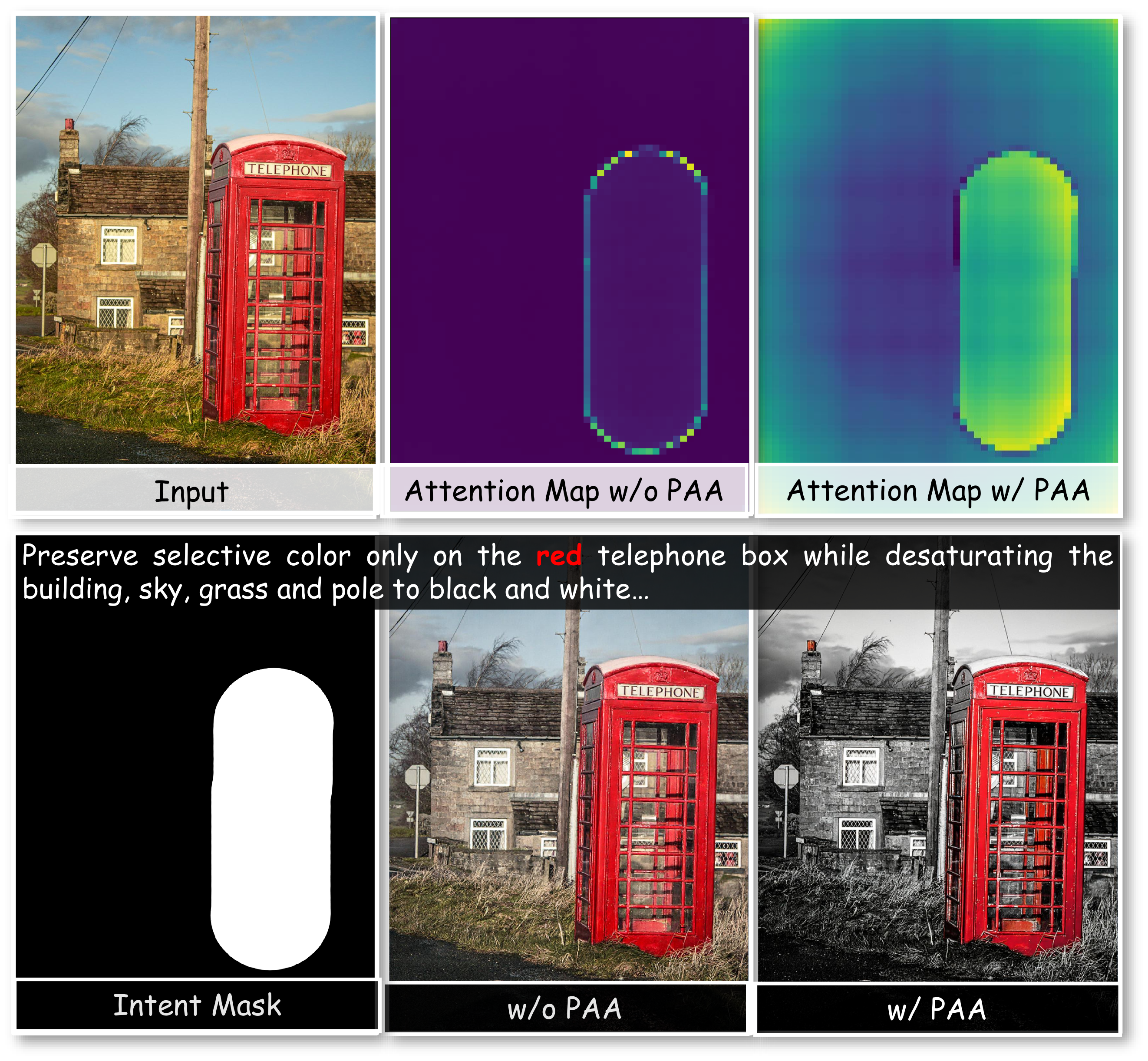}
  \caption{Effect of Pseudo-Intent Guided Attention Alignment.}
    \label{fig:attention_map}
    
  \end{minipage}\hfill
  \begin{minipage}[t]{0.57\linewidth}
    \centering
    \includegraphics[width=\linewidth]{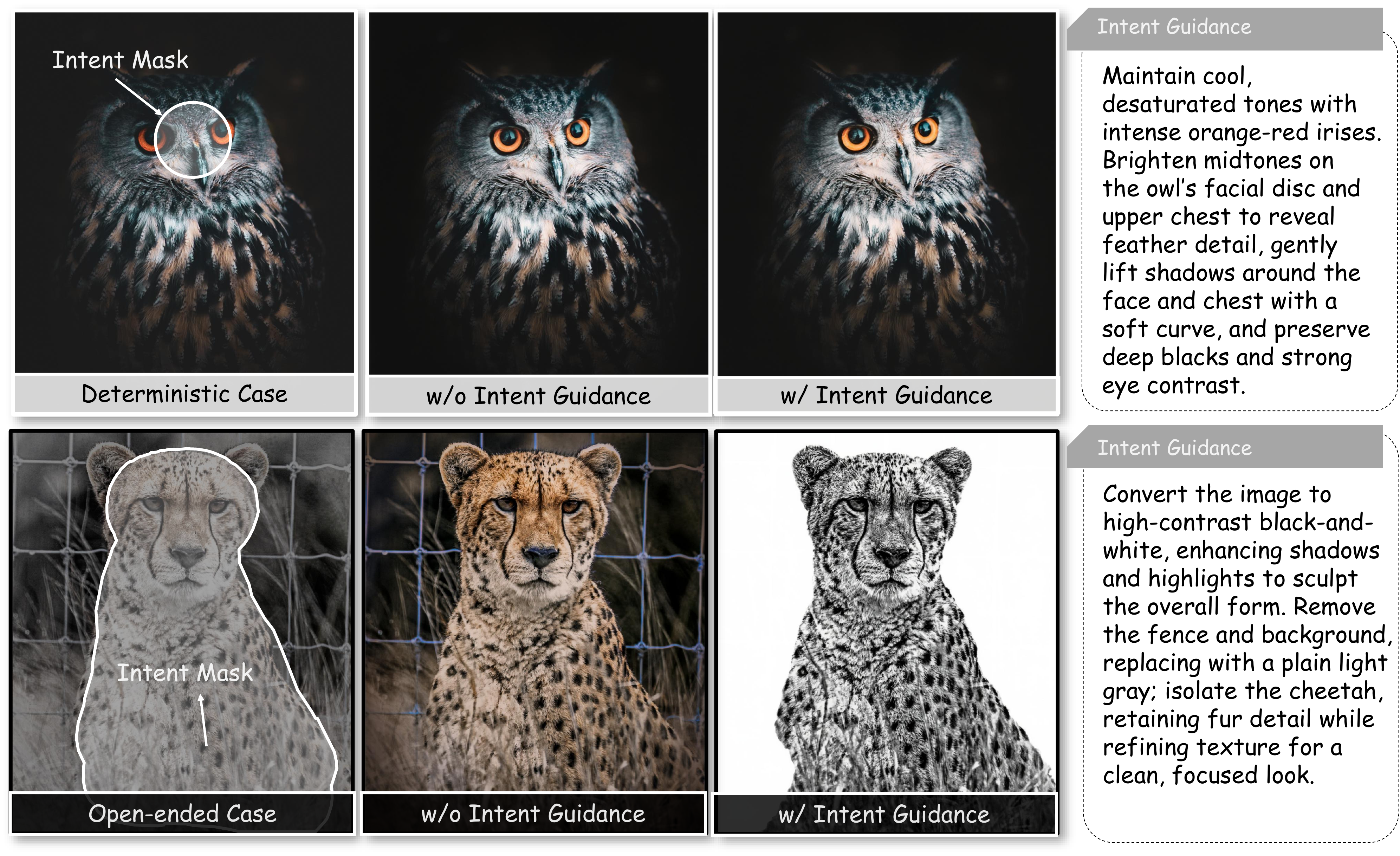}
  \caption{Effect of intent guidance under different levels of user specificity.}
    \label{fig:prompt_ablation}
    
  \end{minipage}
  \label{fig:two_side_by_side}
  \vspace*{-4mm}
\end{figure}
We quantitatively ablate the major components of \methodname in \cref{tab:ablation}. The evaluation is conducted on the full test sets without sampling, covering four representative user scenarios: focus enhancement and multi-round retouching on ControlArt-Bench, global enhancement on MIT-FiveK\cite{five5k}, and local enhancement on MMArt-Bench\cite{lin2025jarvisart}. Starting from a fine-tuned Flux.1-Kontext baseline, adding the planner improves the average performance, suggesting that explicit planning helps constrain the retouching direction. Introducing PAA further improves focus and local editing performance, achieving the best Focus PSNR/SSIM and Local PSNR/SSIM among all variants. This indicates that attention-level supervision is particularly important when the edit needs to redistribute visual prominence within a localized or semantically intended region. Finally, adding operation-consistency loss leads to the best overall average performance across the four scenarios. Although it does not uniformly improve every individual metric, it substantially improves the Global setting, increasing PSNR/SSIM from 17.75/0.7657 to 23.34/0.800, and also yields the best Multi-turn performance. 

We further analyze the contribution of each component in \methodname through focused discussions. We focus our discussion on three primary aspects:

 \vspace*{-4mm}
\subsubsection{Are spatial masks sufficient for visual focus enhancement without explicit attention alignment?} Fig.~\ref{fig:attention_map} suggests that spatial masks alone are insufficient to induce structured focus enhancement. 
In conventional conditioning designs, mask tokens share QKV projections with image conditions and noisy latent tokens. 
However, masks encode intention cues rather than visual appearance. When forced to share projections, attention responses remain diffuse and concentrate along mask boundaries, indicating that the model treats the mask mainly as a spatial constraint rather than a driver of prominence redistribution. Consequently, the edits rely primarily on photometric adjustments, resulting in weak perceptual separation between focal and non-focal regions.

With dedicated mask projections and pseudo-intention attention alignment, the attention distribution becomes more concentrated within the intended region and less boundary-driven. The outputs show clearer structural separation between the preserved red telephone box and the desaturated background. These observations suggest that architectural decoupling and explicit attention supervision are necessary to transform spatial masks into effective mechanisms for visual focus modulation.

 \vspace*{-4mm}
\subsubsection{What role does VLM-driven visual guidance play in focus enhancement?}
\cref{fig:prompt_ablation} further reveals that the impact of VLM-driven guidance depends on the ambiguity of the editing intention. 
When the intended adjustment is explicit and the image offers limited stylistic variation, the difference between using VLM guidance and omitting it is marginal. In such cases, the spatial mask alone provides sufficient structural constraint for reasonable focus enhancement.

However, when multiple plausible adjustment directions exist—such as atmospheric tone balancing, selective desaturation, or contrast redistribution—the absence of VLM guidance leads to inconsistent or unstable edits. Without high-level reasoning, the model may over-enhance local regions or fail to harmonize the global atmosphere with focal emphasis.

These results suggest that VLM-driven guidance mainly acts as a semantic disambiguation mechanism. Rather than amplifying enhancement strength, it constrains the space of plausible editing trajectories, enabling intention-consistent and scene-aware focus modulation in complex retouching scenarios.

\subsubsection{How Does Operation-Consistency Loss Affect Retouching?}
\begin{figure*}[!t]
\vspace{-4mm}
  \centering
  \begin{overpic}[width=\textwidth]{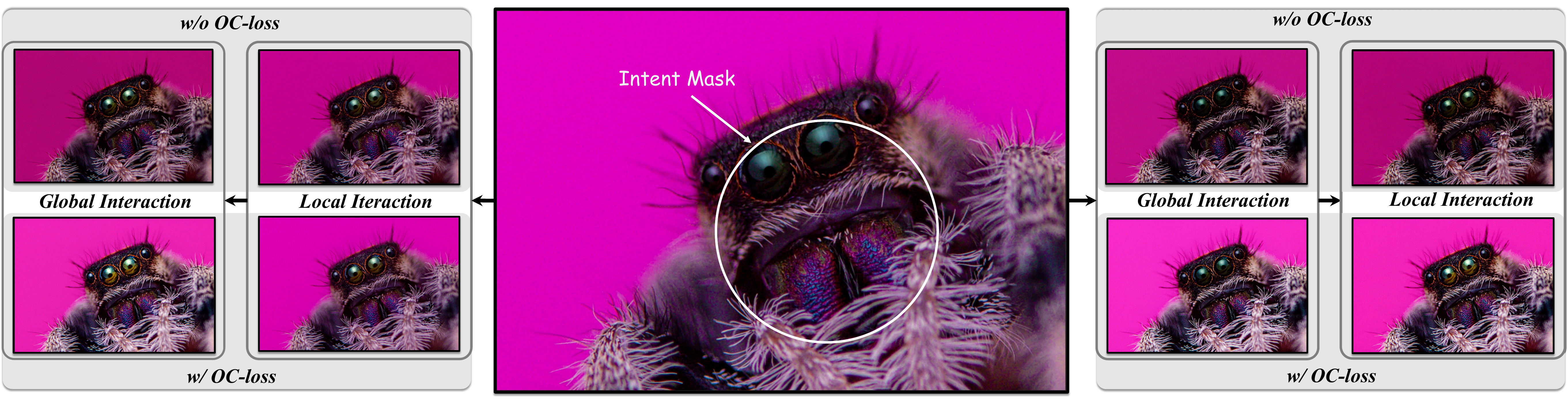}
  \end{overpic}
  \caption{
     Effect of operation-consistency loss (OC-loss) on multi-round retouching.
  }
  \label{fig:con_ablation}
  \vspace{-2mm}
\end{figure*}

\cref{fig:con_ablation} illustrates the effect of operation-consistency regularization on multi-round retouching. 
Without OC-loss, the interaction between global and local adjustments becomes path-dependent: different execution orders (Global→Local vs. Local→Global) produce noticeably different results. In particular, repeated local refinement may override the global atmosphere, while dominant global adjustments can suppress focal emphasis. This suggests that, without structural constraints, heterogeneous editing conditions interfere during diffusion.

With OC-loss enabled, editing trajectories across different execution orders become more consistent. The focal region maintains stable prominence, and the global color tone remains coherent across rounds. This indicates that operation-consistency regularization enforces path-invariant editing dynamics and stabilizes the interaction between global and local directives. Rather than merely aligning pixel outputs, OC-loss reshapes the diffusion process to guide multi-scale adjustments toward a coherent retouching trajectory.

\section{Conclusion}
\vspace{-4mm}
In this work, we revisit image retouching from the perspective of visual focus enhancement, where the goal is not only to improve overall image quality but also to intentionally guide viewers’ attention toward desired regions. To address this problem under weak user intent signals, we propose \methodname, an intent-driven retouching agent that explicitly links user intention, visual focus, and tonal adjustment operations during the retouching process.

The proposed framework combines intention understanding, attention-level focus modulation, and coordinated global–local adjustments to enable controllable visual focus enhancement while preserving perceptual naturalness. In addition, we introduce ControlArt-Bench, a dedicated benchmark with paired spatial intent annotations and focus-oriented evaluation metrics for systematic assessment of visual focus enhancement.

\section{Acknowledgement}
This work is supported in part by the Tianjin Natural Science Foundation Project (25ZXRGGX00290, 24JCJQJC00020, 25JCQNJC01390), the National Natural Science Foundation of China (62306153, 62225604), the Young Elite Scientists Sponsorship Program by CAST (YESS20240686), Shenzhen Science and Technology Program (JCYJ20240813114237048) and the Fundamental Research Funds for the Central Universities (Nankai University, 63253223, 63253219).

\bibliographystyle{splncs04}
\bibliography{main}

@String(CVPR  = {IEEE Conf. Comput. Vis. Pattern Recog.})

@String(AAAI  = {AAAI})

@String(ICPR  = {Int. Conf. Pattern Recog.})

@String(TOG   = {ACM Trans. Graph.})

@String(CVPR  = {CVPR})

@String(ICPR  = {ICPR})

@String(TOG   = {ACM TOG})

@inproceedings{five5k,
  title={Learning photographic global tonal adjustment with a database of input/output image pairs},
  author={Bychkovsky, Vladimir and Paris, Sylvain and Chan, Eric and Durand, Fr{\'e}do},
  booktitle={Proceedings of the IEEE Conference on Computer Vision and Pattern Recognition},
  pages={97--104},
  year={2011},
  organization={IEEE}
}

@inproceedings{liang2021ppr10k,
  title={Ppr10k: A large-scale portrait photo retouching dataset with human-region mask and group-level consistency},
  author={Liang, Jie and Zeng, Hui and Cui, Miaomiao and Xie, Xuansong and Zhang, Lei},
  booktitle={Proceedings of the IEEE/CVF Conference on Computer Vision and Pattern Recognition},
  pages={653--661},
  year={2021}
}

@inproceedings{curve2021starenhancer,
  title={Starenhancer: Learning real-time and style-aware image enhancement},
  author={Song, Yuda and Qian, Hui and Du, Xin},
  booktitle={Proceedings of the IEEE/CVF International Conference on Computer Vision},
  pages={4126--4135},
  year={2021}
}

@inproceedings{duan2025diffretouch,
  title={Diffretouch: Using diffusion to retouch on the shoulder of experts},
  author={Duan, Zheng-Peng and Zhang, Jiawei and Lin, Zheng and Jin, Xin and Wang, XunDong and Zou, Dongqing and Guo, Chun-Le and Li, Chongyi},
  booktitle={Proceedings of the AAAI Conference on Artificial Intelligence},
  volume={39},
  number={3},
  pages={2825--2833},
  year={2025}
}

@inproceedings{li2019hdrnet,
  title={Hdrnet: Single-image-based hdr reconstruction using channel attention cnn},
  author={Li, Jinghui and Fang, Peiyu},
  booktitle={Proceedings of the 2019 4th International Conference on Multimedia Systems and Signal Processing},
  pages={119--124},
  year={2019}
}

@article{gharbi2017deep,
  title={Deep bilateral learning for real-time image enhancement},
  author={Gharbi, Micha{\"e}l and Chen, Jiawen and Barron, Jonathan T and Hasinoff, Samuel W and Durand, Fr{\'e}do},
  journal={ACM Transactions on Graphics (TOG)},
  volume={36},
  number={4},
  pages={1--12},
  year={2017},
  publisher={ACM New York, NY, USA}
}

@inproceedings{3dlut,
  title={Real-time image enhancer via learnable spatial-aware 3d lookup tables},
  author={Wang, Tao and Li, Yong and Peng, Jingyang and Ma, Yipeng and Wang, Xian and Song, Fenglong and Yan, Youliang},
  booktitle={Proceedings of the IEEE/CVF International Conference on Computer Vision},
  pages={2471--2480},
  year={2021}
}

@inproceedings{yang2022adaint,
  title={AdaInt: Learning adaptive intervals for 3D lookup tables on real-time image enhancement},
  author={Yang, Canqian and Jin, Meiguang and Jia, Xu and Xu, Yi and Chen, Ying},
  booktitle={Proceedings of the IEEE/CVF Conference on Computer Vision and Pattern Recognition},
  pages={17522--17531},
  year={2022}
}

@article{lut2,
  title={Learning image-adaptive 3d lookup tables for high performance photo enhancement in real-time},
  author={Zeng, Hui and Cai, Jianrui and Li, Lida and Cao, Zisheng and Zhang, Lei},
  journal={IEEE Transactions on Pattern Analysis and Machine Intelligence},
  volume={44},
  number={4},
  pages={2058--2073},
  year={2020},
  publisher={IEEE}
}

@misc{curve2022cudi,
      title={CuDi: Curve Distillation for Efficient and Controllable Exposure Adjustment}, 
      author={Chongyi Li and Chunle Guo and Ruicheng Feng and Shangchen Zhou and Chen Change Loy},
      year={2022},
      eprint={2207.14273},
      archivePrefix={arXiv},
      primaryClass={cs.CV},
      url={https://arxiv.org/abs/2207.14273}, 
}

@article{curve2021zerodce,
  title={Learning to enhance low-light image via zero-reference deep curve estimation},
  author={Li, Chongyi and Guo, Chunle and Loy, Chen Change},
  journal={IEEE transactions on pattern analysis and machine intelligence},
  volume={44},
  number={8},
  pages={4225--4238},
  year={2021},
  publisher={IEEE}
}

@inproceedings{curve2021curl,
  title={Curl: Neural curve layers for global image enhancement},
  author={Moran, Sean and McDonagh, Steven and Slabaugh, Gregory},
  booktitle={2020 25th International Conference on Pattern Recognition (ICPR)},
  pages={9796--9803},
  year={2021},
  organization={IEEE}
}

@inproceedings{retargeting2011wacv,
  title={Saliency retargeting: An approach to enhance image aesthetics},
  author={Wong, Lai-Kuan and Low, Kok-Lim},
  booktitle={2011 IEEE Workshop on Applications of Computer Vision (WACV)},
  pages={73--80},
  year={2011},
  organization={IEEE}
}

@inproceedings{retargeting2023realistic,
  title={Realistic saliency guided image enhancement},
  author={Miangoleh, S Mahdi H and Bylinskii, Zoya and Kee, Eric and Shechtman, Eli and Aksoy, Ya{\u{g}}iz},
  booktitle={Proceedings of the IEEE/CVF Conference on Computer Vision and Pattern Recognition},
  pages={186--194},
  year={2023}
}

@inproceedings{reducing2022deep,
  title={Deep saliency prior for reducing visual distraction},
  author={Aberman, Kfir and He, Junfeng and Gandelsman, Yossi and Mosseri, Inbar and Jacobs, David E and Kohlhoff, Kai and Pritch, Yael and Rubinstein, Michael},
  booktitle={Proceedings of the IEEE/CVF Conference on Computer Vision and Pattern Recognition},
  pages={19851--19860},
  year={2022}
}

@inproceedings{
lin2025jarvisart,
title={JarvisArt: Liberating Human Artistic Creativity via an Intelligent Photo Retouching Agent},
author={Yunlong Lin and ZiXu Lin and Kunjie Lin and Jinbin Bai and Panwang Pan and Chenxin Li and Haoyu Chen and Zhongdao Wang and Xinghao Ding and Wenbo Li and Shuicheng YAN},
booktitle={The Thirty-ninth Annual Conference on Neural Information Processing Systems},
year={2025}
}

@inproceedings{chang2026pertouch,
    title     = {PerTouch: A Unified Diffusion-based Image Retouching Framework with VLM-driven Agent},
    author    = {Chang, Zewei and Duan, Zheng-Peng and Zhang, Jianxing and others},
    year      = 2026,
    booktitle = {The 40th Annual AAAI Conference on Artificial Intelligence}
}

@misc{lin2025jarvisevoselfevolvingphotoediting,
      title={JarvisEvo: Towards a Self-Evolving Photo Editing Agent with Synergistic Editor-Evaluator Optimization}, 
      author={Yunlong Lin and Linqing Wang and Kunjie Lin and Zixu Lin and Kaixiong Gong and Wenbo Li and Bin Lin and Zhenxi Li and Shiyi Zhang and Yuyang Peng and Wenxun Dai and Xinghao Ding and Chunyu Wang and Qinglin Lu},
      year={2025},
      eprint={2511.23002},
      archivePrefix={arXiv},
      primaryClass={cs.CV},
      url={https://arxiv.org/abs/2511.23002}, 
}

@misc{li2025uniworldv2,
      title={Uniworld-V2: Reinforce Image Editing with Diffusion Negative-aware Finetuning and MLLM Implicit Feedback}, 
      author={Zongjian Li and Zheyuan Liu and Qihui Zhang and Bin Lin and Feize Wu and Shenghai Yuan and Zhiyuan Yan and Yang Ye and Wangbo Yu and Yuwei Niu and Shaodong Wang and Xinhua Cheng and Li Yuan},
      year={2025},
      eprint={2510.16888},
      archivePrefix={arXiv},
      primaryClass={cs.CV},
      url={https://arxiv.org/abs/2510.16888}, 
}

@misc{stepedit,
      title={ReasonEdit: Towards Reasoning-Enhanced Image Editing Models}, 
      author={Fukun Yin and Shiyu Liu and Yucheng Han and Zhibo Wang and Peng Xing and Rui Wang and Wei Cheng and Yingming Wang and Aojie Li and Zixin Yin and Pengtao Chen and Xiangyu Zhang and Daxin Jiang and Xianfang Zeng and Gang Yu},
      year={2025},
      eprint={2511.22625},
      archivePrefix={arXiv},
      primaryClass={cs.CV},
      url={https://arxiv.org/abs/2511.22625}, 
}

@misc{labs2025flux1kontextflowmatching,
      title={FLUX.1 Kontext: Flow Matching for In-Context Image Generation and Editing in Latent Space}, 
      author={Black Forest Labs and Stephen Batifol and Andreas Blattmann and Frederic Boesel and Saksham Consul and Cyril Diagne and Tim Dockhorn and Jack English and Zion English and Patrick Esser and Sumith Kulal and Kyle Lacey and Yam Levi and Cheng Li and Dominik Lorenz and Jonas Müller and Dustin Podell and Robin Rombach and Harry Saini and Axel Sauer and Luke Smith},
      year={2025},
      eprint={2506.15742},
      archivePrefix={arXiv},
      primaryClass={cs.GR},
      url={https://arxiv.org/abs/2506.15742}, 
}

@misc{wu2025qwenimagetechnicalreport,
      title={Qwen-Image Technical Report}, 
      author={Chenfei Wu and Jiahao Li and Jingren Zhou and Junyang Lin and Kaiyuan Gao and Kun Yan and Sheng-ming Yin and Shuai Bai and Xiao Xu and Yilei Chen and Yuxiang Chen and Zecheng Tang and Zekai Zhang and Zhengyi Wang and An Yang and Bowen Yu and Chen Cheng and Dayiheng Liu and Deqing Li and Hang Zhang and Hao Meng and Hu Wei and Jingyuan Ni and Kai Chen and Kuan Cao and Liang Peng and Lin Qu and Minggang Wu and Peng Wang and Shuting Yu and Tingkun Wen and Wensen Feng and Xiaoxiao Xu and Yi Wang and Yichang Zhang and Yongqiang Zhu and Yujia Wu and Yuxuan Cai and Zenan Liu},
      year={2025},
      eprint={2508.02324},
      archivePrefix={arXiv},
      primaryClass={cs.CV},
      url={https://arxiv.org/abs/2508.02324}, 
}

@misc{geminiteam2025geminifamilyhighlycapable,
      title={Gemini: A Family of Highly Capable Multimodal Models}, 
      author={Gemini Team and Rohan Anil and Sebastian Borgeaud and Jean-Baptiste Alayrac and Jiahui Yu and Radu Soricut and Johan Schalkwyk and Andrew M. Dai and Anja Hauth and Katie Millican and David Silver and Melvin Johnson and others},
      year={2025},
      eprint={2312.11805},
      archivePrefix={arXiv},
      primaryClass={cs.CL},
      url={https://arxiv.org/abs/2312.11805}, 
}

@inproceedings{kummererSaliencyBenchmarkingMade2018,
    title = {Saliency Benchmarking Made Easy: Separating Models, Maps and Metrics},
    pages = {798--814},
    booktitle = {European Conference on Computer Vision},
    author = {K{\"u}mmerer, Matthias and Wallis, Thomas S. A. and Bethge, Matthias},
    year = {2018},
}

@misc{bai2025qwen25vltechnicalreport,
      title={Qwen2.5-VL Technical Report}, 
      author={Shuai Bai and Keqin Chen and Xuejing Liu and Jialin Wang and Wenbin Ge and Sibo Song and Kai Dang and Peng Wang and Shijie Wang and Jun Tang and Humen Zhong and Yuanzhi Zhu and Mingkun Yang and Zhaohai Li and Jianqiang Wan and Pengfei Wang and Wei Ding and Zheren Fu and Yiheng Xu and Jiabo Ye and Xi Zhang and Tianbao Xie and Zesen Cheng and Hang Zhang and Zhibo Yang and Haiyang Xu and Junyang Lin},
      year={2025},
      eprint={2502.13923},
      archivePrefix={arXiv},
      primaryClass={cs.CV},
      url={https://arxiv.org/abs/2502.13923}, 
}

@inproceedings{chen2018deep,
  title={Deep photo enhancer: Unpaired learning for image enhancement from photographs with gans},
  author={Chen, Yu-Sheng and Wang, Yu-Ching and Kao, Man-Hsin and Chuang, Yung-Yu},
  booktitle={Proceedings of the IEEE Conference on Computer Vision and Pattern Recognition},
  pages={6306--6314},
  year={2018}
}

@INPROCEEDINGS{saliencyguided,
  author={Jiang, Lai and Xu, Mai and Wang, Xiaofei and Sigal, Leonid},
  booktitle={2021 IEEE/CVF Conference on Computer Vision and Pattern Recognition (CVPR)}, 
  title={Saliency-Guided Image Translation}, 
  year={2021},
  volume={},
  number={},
  pages={16504-16513},
}

@article{dutt2025monetgpt,
  title={MonetGPT: Solving Puzzles Enhances MLLMs' Image Retouching Skills},
  author={Dutt, Niladri Shekhar and Ceylan, Duygu and Mitra, Niloy J},
  journal={ACM Transactions on Graphics (TOG)},
  volume={44},
  number={4},
  pages={1--12},
  year={2025},
  publisher={ACM New York, NY, USA}
}

@misc{wu2026retouchiq,
      title={RetouchIQ: MLLM Agents for Instruction-Based Image Retouching with Generalist Reward}, 
      author={Qiucheng Wu and Jing Shi and Simon Jenni and Kushal Kafle and Tianyu Wang and Shiyu Chang and Handong Zhao},
      year={2026},
      eprint={2602.17558},
      archivePrefix={arXiv},
      primaryClass={cs.CV},
      url={https://arxiv.org/abs/2602.17558}, 
}

@inproceedings{
ku2024imagenhub,
title={ImagenHub: Standardizing the evaluation of conditional image generation models},
author={Max Ku and Tianle Li and Kai Zhang and Yujie Lu and Xingyu Fu and Wenwen Zhuang and Wenhu Chen},
booktitle={The Twelfth International Conference on Learning Representations},
year={2024}
}

@misc{openai2024gptimage,
  title        = {GPT-Image-1.5},
  author       = {{OpenAI}},
  year         = {2025},
  howpublished = {\url{https://openai.com/zh-Hans-CN/index/new-chatgpt-images-is-here/}},
  note         = {Accessed: 2026}
}

@inproceedings{wen2024retouchformer,
  title={Retouchformer: Semi-supervised high-quality face retouching transformer with prior-based selective self-attention},
  author={Wen, Xue and Xie, Lianxin and Jiang, Le and Chen, Tianyi and Wu, Si and Liu, Cheng and Wong, Hau-San},
  booktitle={Proceedings of the AAAI Conference on Artificial Intelligence},
  volume={38},
  number={6},
  pages={5903--5911},
  year={2024}
}

@misc{yang2025qwen3technicalreport,
      title={Qwen3 Technical Report}, 
      author={An Yang and Anfeng Li and Baosong Yang and Beichen Zhang and Binyuan Hui and Bo Zheng and Bowen Yu and Chang Gao and Chengen Huang and Chenxu Lv and Chujie Zheng and Dayiheng Liu and Fan Zhou and Fei Huang and Feng Hu and Hao Ge and Haoran Wei and Huan Lin and Jialong Tang and Jian Yang and Jianhong Tu and Jianwei Zhang and Jianxin Yang and Jiaxi Yang and Jing Zhou and Jingren Zhou and Junyang Lin and Kai Dang and Keqin Bao and Kexin Yang and Le Yu and Lianghao Deng and Mei Li and Mingfeng Xue and Mingze Li and Pei Zhang and Peng Wang and Qin Zhu and Rui Men and Ruize Gao and Shixuan Liu and Shuang Luo and Tianhao Li and Tianyi Tang and Wenbiao Yin and Xingzhang Ren and Xinyu Wang and Xinyu Zhang and Xuancheng Ren and Yang Fan and Yang Su and Yichang Zhang and Yinger Zhang and Yu Wan and Yuqiong Liu and Zekun Wang and Zeyu Cui and Zhenru Zhang and Zhipeng Zhou and Zihan Qiu},
      year={2025},
      eprint={2505.09388},
      archivePrefix={arXiv},
      primaryClass={cs.CL},
      url={https://arxiv.org/abs/2505.09388}, 
}

@InProceedings{saliencytransformer,
    author    = {Liu, Nian and Zhang, Ni and Wan, Kaiyuan and Shao, Ling and Han, Junwei},
    title     = {Visual Saliency Transformer},
    booktitle = {Proceedings of the IEEE/CVF International Conference on Computer Vision},
    month     = {October},
    year      = {2021},
    pages     = {4722-4732}
}

@ARTICLE{saliency1998,
  author={Itti, L. and Koch, C. and Niebur, E.},
  journal={IEEE Transactions on Pattern Analysis and Machine Intelligence}, 
  title={A model of saliency-based visual attention for rapid scene analysis}, 
  year={1998},
  volume={20},
  number={11},
  pages={1254-1259},
}

@inproceedings{kim2020pienet,
  title={PieNet: Personalized image enhancement network},
  author={Kim, Han-Ul and Koh, Young Jun and Kim, Chang-Su},
  booktitle={European Conference on Computer Vision},
  pages={374--390},
  year={2020},
  organization={Springer}
}

\clearpage
\section*{\Large{Appendices}}
\appendix

Our supplemental materials include the following parts:
\begin{itemize}

\item \cref{sec:train} Training Details
\begin{itemize}
    \item \cref{sec:train_set} Composition of the Training Set.
    \item \cref{sec:train_prompt} Intent Guidance during Training.
    \item \cref{sec:train_mode} Inputs for Retouching Executor.
\end{itemize}

\item \cref{sec:metrics} Details of Metrics
\begin{itemize}
    \item \cref{sec:metrics_limit} Limitations of Existing Metrics.
    \item \cref{sec:metrics_fa} Metrics for Visual Focus Enhancement.
    \item \cref{sec:validity} Metric Validity.
    \item \cref{sec:metrics_prompt} Prompts for MLLM-Based Metrics.
\end{itemize}
\item \cref{sec:add_exp} Additional Experimental Results.
\begin{itemize}
    \item \cref{sec:gir} Comparison on Global Image Retouching.
    \item \cref{sec:lir} Comparison on Local Image Retouching.
    \item \cref{sec:corseness} Comparison on Local Image Retouching.
    \item \cref{sec:fir} More Visual Comparisons.
    \item \cref{sec:more_instruction} Instruction for Compared Methods
\end{itemize}
\item \cref{sec:limit} Limitations.
\end{itemize}

\section{Training Details}
\label{sec:train}
\subsection{Composition of the Training Set}
\label{sec:train_set}
\begin{figure*}[h]
\vspace{-4mm}
  \centering
  \begin{overpic}[width=\textwidth]{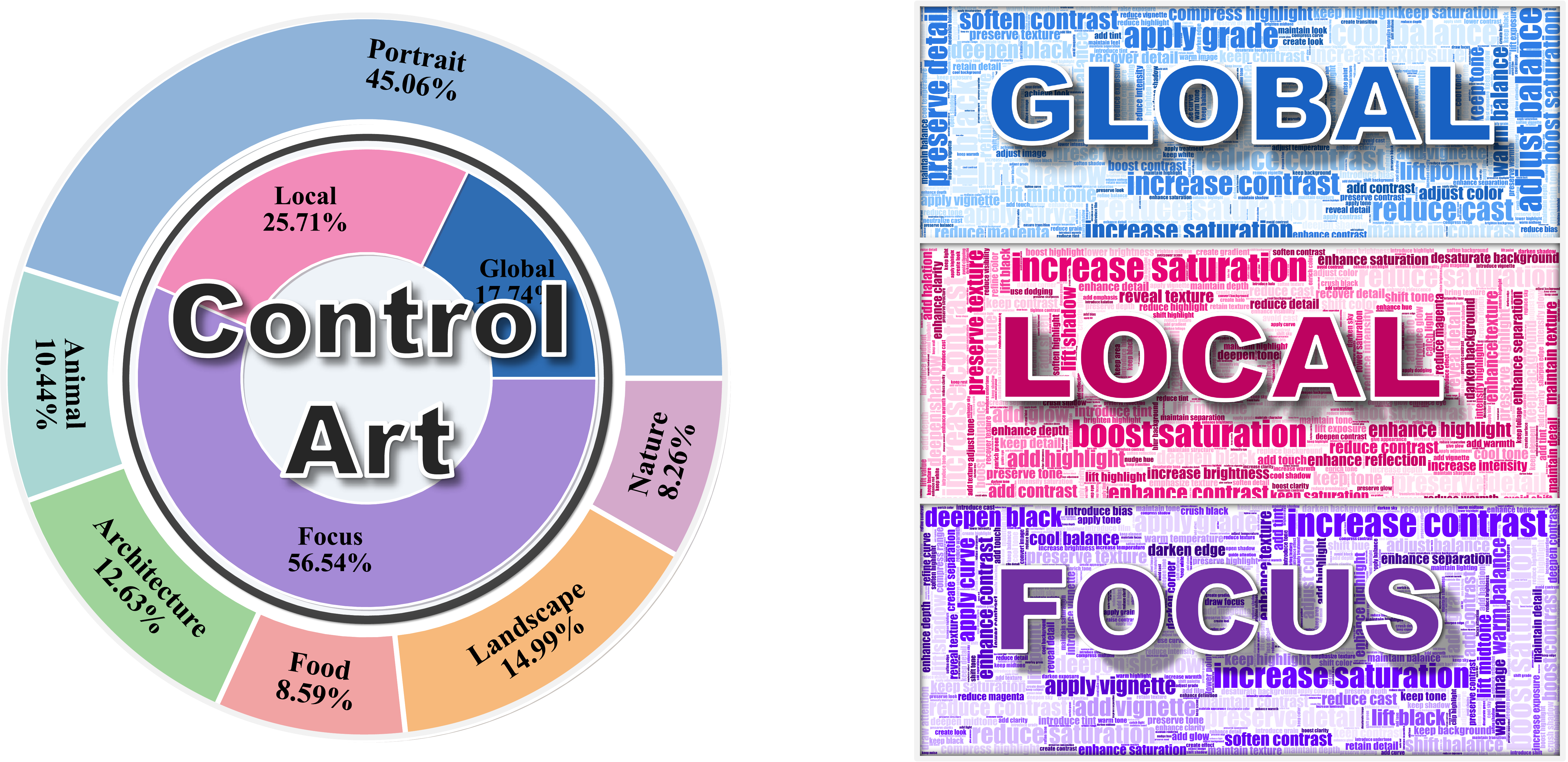}
    \scriptsize
    \put(22,-2){(a)}
    \put(77,-2){(b)}
  \end{overpic}
  \caption{
Overview of ControlArt. (a) Category and editing-mode composition of the dataset: the outer ring shows semantic categories, while the inner ring summarizes edit modes (Global/Local/Focus). (b) Word clouds of instruction keywords for Global, Local, and Focus modes, highlighting mode-specific editing patterns and common retouching operations.
  }
  \label{fig:dataset}
\end{figure*}

The statistics of our training set (we called ControlArt) are shown in \cref{fig:dataset}. 
The dataset consists of three types of samples corresponding to the execution modes in our framework: 
(1) \textbf{focus retouching data}, 
(2) \textbf{global retouching data}, and 
(3) \textbf{local retouching data}, 
which are used to train the \emph{Focus Mode}, \emph{Global Mode}, and \emph{Local Mode}, respectively.

For category (1), we follow the data generation pipeline described in the main text to collect retouching pairs, extract user intention regions, and construct the final \textit{Intent Mask}. 
To strengthen the model’s capability in global adjustments, local refinements, and their coordination, we additionally incorporate samples from global and local retouching datasets.

For global retouching data, we filter samples from PPR10k\cite{liang2021ppr10k} by removing pairs that introduce noticeable saliency shifts, retaining only those with global adjustments but no explicit visual focus enhancement. 

For local retouching data, we select pairs containing only local editing operations (i.e., mask-based adjustments). Based on the corresponding operations, we obtain high-quality \textit{Intent Masks} through a combination of automatic generation and manual annotation to train the \emph{Local Mode}.
\subsection{Intent Guidance during Training}
\label{sec:train_prompt}
During training, we use the pre- and post-retouch images $I$ and $I_{gt}$ to derive intent guidance $\mathcal{G}$ for each sample. Specifically, we extract scene-level descriptions from $I$ and $I_{gt}$, obtaining $d_s$ and $d^{gt}_s$. By comparing the paired representations $\langle I, d_s \rangle$ and $\langle I_{gt}, d^{gt}_s \rangle$, we infer the user intention $d_u$. The subsequent steps follow the pipeline described in the main text: the inferred intention is used to generate the initial guidance $\mathcal{G}_0$, which is further formatted into a global edit instruction $\mathcal{G}_{\text{global}}$ and a local edit instruction $\mathcal{G}_{\text{local}}$.

For training samples corresponding to different modes, the intent guidance is explicitly organized according to the target mode to form the actual model input. The detailed training input formats for different modes are provided in \cref{sec:train_mode}.
\subsection{Inputs for Retouching Executor}
\label{sec:train_mode}
As described in the main text, different modes use different input formats.
Here, we provide a more detailed description of how the inputs to the retouching executor are organized during training for different types of samples:
\begin{itemize}
\item Global Mode. We use global retouching data to train. The input of the retouching executor is $\langle I, \mathcal{G}_{global}, M_0 \rangle$ where $M_0$ is an all-zero mask so that the retouching is driven purely by $\mathcal{G}_{global}$.

\item Local Mode. We use local retouching data to train. $\langle I, \mathcal{G}_{local}, M_u \rangle$ is the input.

\item Focus Mode. We use visual focus enhancement retouching data to train. $\langle I,\mathcal{G}_{global} + \mathcal{G}_{local}, M_u \rangle$ is the input.

\end{itemize}

\section{Details Of Metrics}
\label{sec:metrics}
\begin{figure*}[!h
]
\vspace{-4mm}
  \centering
  \begin{overpic}[width=\textwidth]{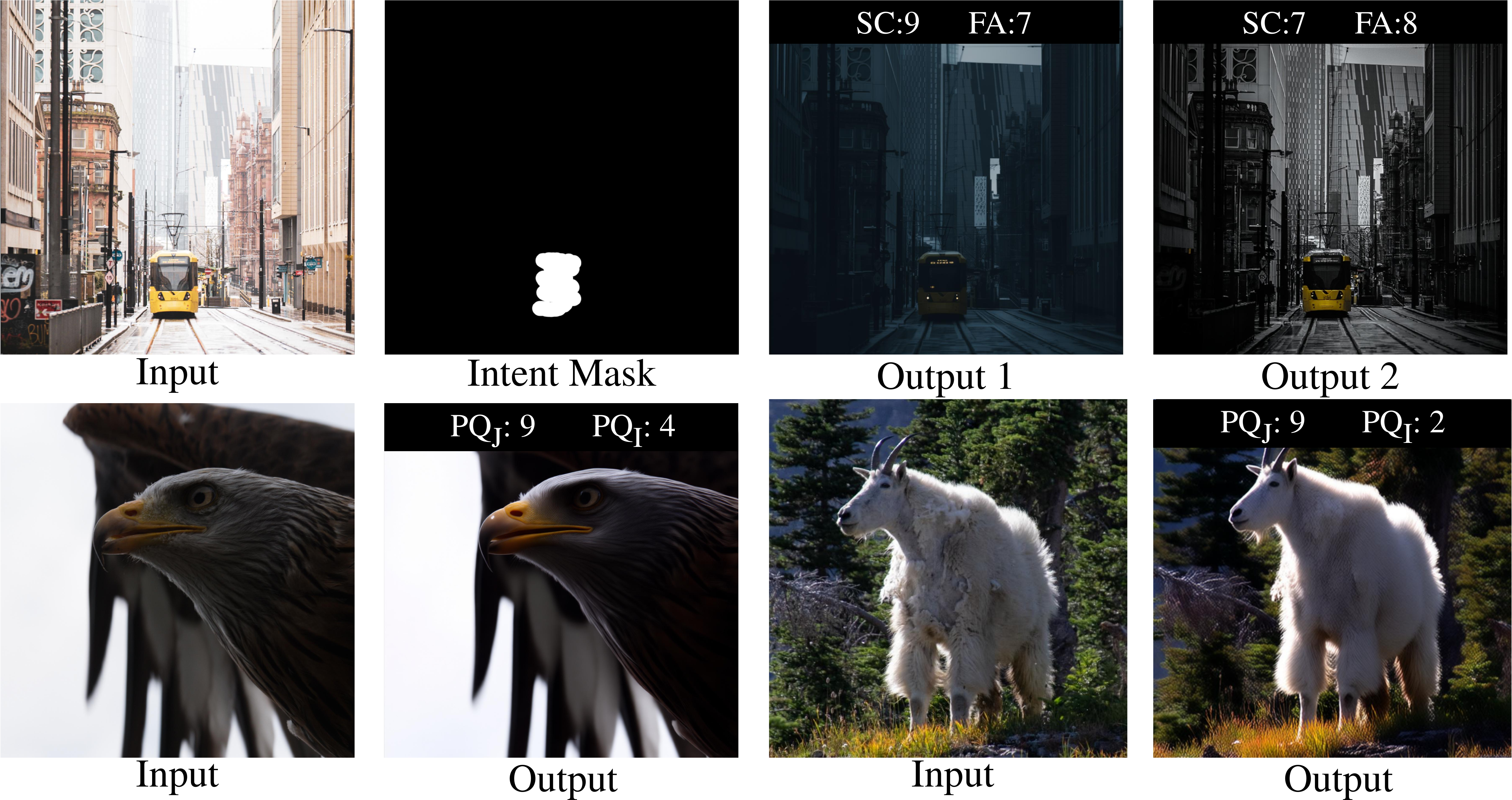}
  \end{overpic}
  \caption{
     Top: Compared with SC, FA better reflects whether the visual focus aligns with the user intent.
Bottom: $\text{PQ}_{\text{I}}$, which evaluates perceptual quality independently, provides a more reliable measure of content consistency and artifacts than $\text{PQ}_{\text{J}}$, which predicts perceptual quality jointly with SC.
  }
  \label{fig:metrics}
\end{figure*}
\vspace{-9mm}
\subsection{Limitations of Existing Metrics}
\label{sec:metrics_limit}
\cref{fig:metrics} exposes two limitations of existing metrics for visual focus enhancement, particularly SC and PQ\cite{ku2024imagenhub}.

First, the commonly used SC metric does not explicitly measure whether the retouched image shifts visual attention toward the user-specified intent region. A higher SC score does not necessarily imply better alignment between the retouched result and the intended focus.

Second, existing VLM-based perceptual-quality evaluation can be affected by the way the metric is predicted. 
In our analysis, we distinguish two variants of Perceptual Quality (PQ): $\text{PQ}_{\text{J}}$ and $\text{PQ}_{\text{I}}$. 
Here, $\text{PQ}_{\text{J}}$ denotes jointly predicted perceptual quality, where the VLM evaluates PQ together with SC in a single judgement (\cite{ku2024imagenhub} evaluates as this way). 
Although this setting is convenient, the PQ score may be influenced by other evaluation dimensions. 
In contrast, $\text{PQ}_{\text{I}}$ denotes independently predicted perceptual quality, where the VLM is asked to evaluate only content preservation and visual artifacts, without jointly considering other metrics. 
This independent formulation reduces cross-metric interference and provides a cleaner estimate of content fidelity. 
As shown in the bottom row of Fig.~\ref{fig:metrics}, $\text{PQ}_{\text{I}}$ better reflects content consistency and visual artifacts than $\text{PQ}_{\text{J}}$.

\subsection{Metrics for Visual Focus Enhancement}
\label{sec:metrics_fa}
To address the above limitation, we introduce \textbf{Focus Alignment (FA)}, a metric designed to evaluate whether the visual focus of the retouched image is consistent with the user-intended focus region. FA directly measures the alignment between the retouched visual focus and the target region specified by the user, and is scored on a 0–10 scale. The evaluation considers: (1) the primary visual landing point of the viewer; (2) the clarity of the overall visual hierarchy; (3) the enhancement of the target region through contrast or saliency; (4) the use of attention-guidance mechanisms such as lighting, tonal shaping, or vignetting; and (5) the stability of the target region as the dominant perceptual anchor. In this way, FA complements existing metrics by capturing a core property of visual focus enhancement that SC does not directly reflect.
In addition, we compute PQ independently rather than jointly with other metrics, as this provides a more reliable assessment of content consistency and artifact severity.

To ensure reproducibility, we use Qwen2.5-VL-72B\cite{bai2025qwen25vltechnicalreport} as the evaluator throughout all experiments. Although we observe that MLLM-based evaluation is not always fully accurate, it still provides a relatively consistent and reproducible protocol for large-scale assessment. We hope future evaluation methods can better align with human perception and user intent.
\subsection{Metric Validity}
\label{sec:validity}
To further validate whether FA and $\text{PQ}_{\text{I}}$ better agree with human judgement, we compare VLM-based metrics with human preferences on 2,000 user-study pairs. 
We evaluate two human-annotated dimensions: Intent Consistency (IC), which measures whether the retouched result follows the user intent, and Content Fidelity (CF), which measures whether the result preserves image content and avoids artifacts. 
For IC, we compare SC and FA. 
For CF, we compare $\text{PQ}_{\text{J}}$ and $\text{PQ}_{\text{I}}$.

\begin{table}[t]
\centering
\small
\setlength{\tabcolsep}{1pt}
\renewcommand{\arraystretch}{0.9}
\caption{
Human-VLM preference correlation validation on metric validity (left) and quantitative ablation on intent coarseness level (right). }
\begin{minipage}[t]{0.58\columnwidth}
\centering
\begin{adjustbox}{max width=\linewidth}
\begin{tabular}{l|cc|cc}
\toprule
Metrics Correlation & IC-SC & IC-FA & CF-$\text{PQ}_{\text{J}}$ & CF-$\text{PQ}_{\text{I}}$ \\
\midrule
Spearman $\rho$ & 0.260 & \q{0.370} & 0.331 & \q{0.390} \\
Kendall $\tau$ & 0.236 & \q{0.348} & 0.301 & \q{0.357} \\
Human tie rate & 10.6\% & 10.6\% & 36.6\% & 36.6\% \\
VLM tie rate & 40.4\% & 74.5\% & 41.3\% & 56.1\% \\
Winner judgement acc. & 67.9\% & \q{87.6\%} & 74.6\% & \q{83.42\%} \\
\bottomrule
\end{tabular}
\end{adjustbox}
\end{minipage}
\hspace{0.01\columnwidth}
\begin{minipage}[t]{0.39\columnwidth}
\centering
\setlength{\tabcolsep}{1pt}
\begin{adjustbox}{max width=\linewidth}
\begin{tabular}{l|cccc}
\toprule
Mask Type & PSNR & SSIM & O & KL \\
\midrule
Mixed & 21.88 & 0.8517 & 8.71 & 0.2215 \\
Click & 21.95 & 0.8563 & 8.65 & 0.2005 \\
Strike & 22.08 & 0.8566 & 8.66 & 0.2144 \\
Region & 21.84 & 0.8515 & 8.66 & 0.2280 \\
SAM & 21.66 & 0.8432 & 8.62 & 0.2407 \\
\bottomrule
\end{tabular}
\end{adjustbox}
\end{minipage}

\label{tab:metric_and_coarseness}
\end{table}

As shown in Tab.~\ref{tab:metric_and_coarseness} left, FA is more consistent with human-evaluated Intent Consistency than SC. 
It improves Spearman correlation from $0.260$ to $0.370$, Kendall correlation from $0.236$ to $0.348$, and winner judgement accuracy from $67.9\%$ to $87.6\%$. 
These results support our claim that FA better measures whether the retouched visual focus follows the user intent.

For Content Fidelity, $\text{PQ}_{\text{I}}$ also consistently outperforms $\text{PQ}_{\text{J}}$. 
It improves Spearman correlation from $0.331$ to $0.390$, Kendall correlation from $0.301$ to $0.357$, and winner judgement accuracy from $74.6\%$ to $83.42\%$. 
This confirms that independently evaluating perceptual quality reduces interference from other judgement dimensions and provides a more reliable measure of content preservation and visual artifacts.

We also observe that VLM-based metrics are generally more conservative than human annotators and are more likely to predict ties. 
For example, the VLM tie rate of IC-FA is $74.5\%$, while the human tie rate is only $10.6\%$. 
This conservativeness explains why the average automatic metric scores in the main paper can appear close, whereas the user study reveals more pronounced pairwise preferences. 
Therefore, FA, $\text{PQ}_{\text{I}}$, and O are meaningful automatic metrics for large-scale evaluation, but they should be viewed as complementary to human studies rather than replacements for them.
\subsection{Prompts for MLLM-Based Metrics}
\label{sec:metrics_prompt}
\cref{fig:prompt_metrics} presents the prompts used to evaluate Focus Alignment (FA) and Perceptual Quality (PQ). In the local retouching setting, we use the foreground mask to isolate the portrait region and evaluate all metrics on the resulting masked portrait image. Accordingly, we denote the perceptual quality metric in this setting as PQ$^{\mathrm{RC}}$. The instance-level overall score is computed as the geometric mean of FA and PQ, \ie, $O = \sqrt{\text{FA} \times \text{PQ}}$.
\begin{figure*}[!t]
\vspace{-4mm}
  \centering
  \begin{overpic}[width=\textwidth]{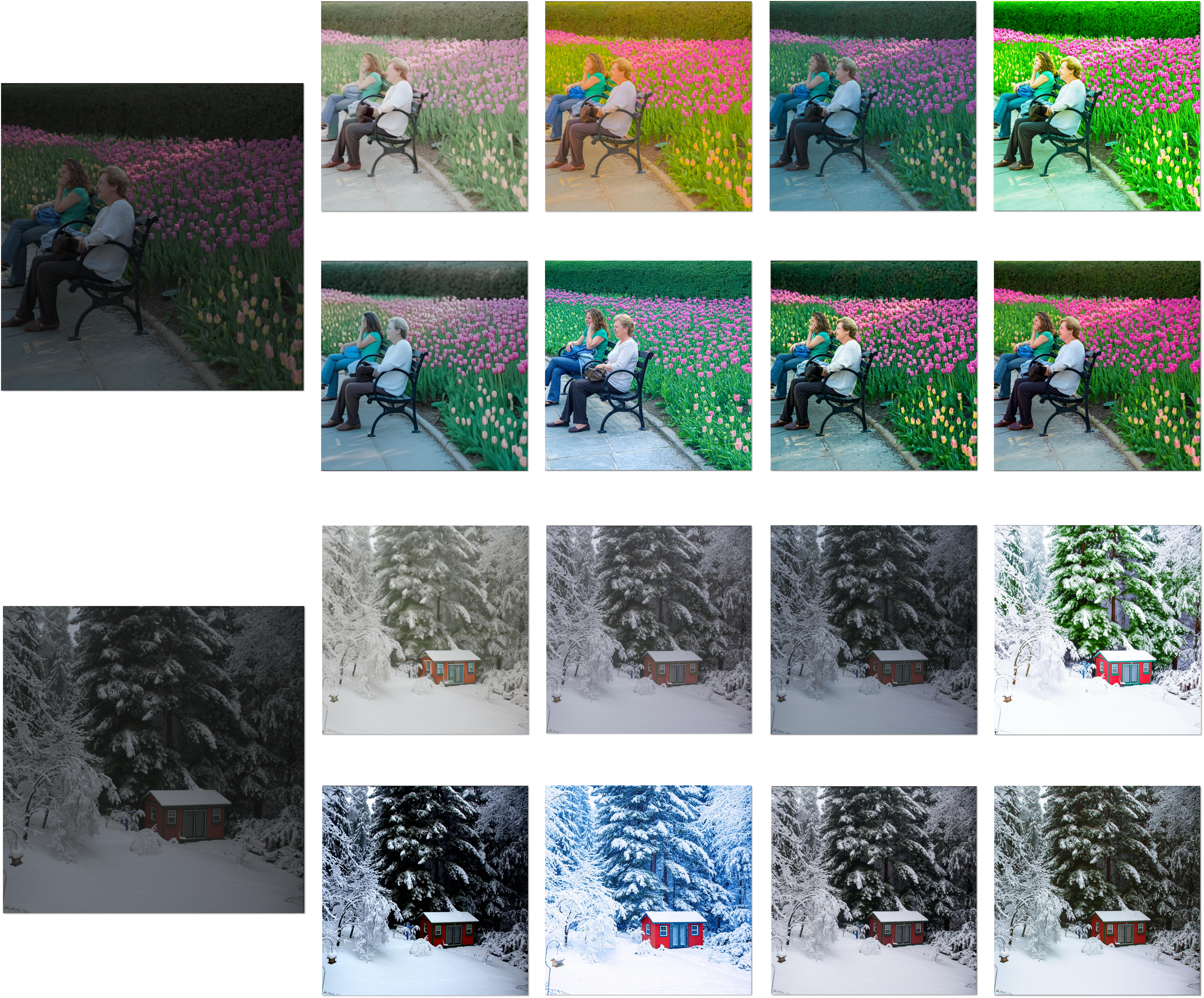}
  \put(10,4){Input}
  \put(30,19){Pertouch}
  \put(48,19){JarvisArt}
  \put(66,19){JarvisEvo}
  \put(84,19){UniWorld-v2}
  \put(27.5,-3){Step1X-Edit}
  \put(44.5,-3){GPT-Image-1.5}
  \put(63.5,-3){Nano-Banana 2}
  \put(87,-3){Ours}
  \put(10,48){Input}
  \put(30,63){Pertouch}
  \put(48,63){JarvisArt}
  \put(66,63){JarvisEvo}
  \put(84,63){UniWorld-v2}
  \put(27.5,41){Step1X-Edit}
  \put(44.5,41){GPT-Image-1.5}
  \put(63.5,41){Nano-Banana 2}
  \put(87,41){Ours}
  
  \end{overpic}
  \vspace{1mm}
  \caption{
     Visual comparison on MIT-Adobe FiveK\cite{five5k}.
  }
  \label{fig:global}
\end{figure*}
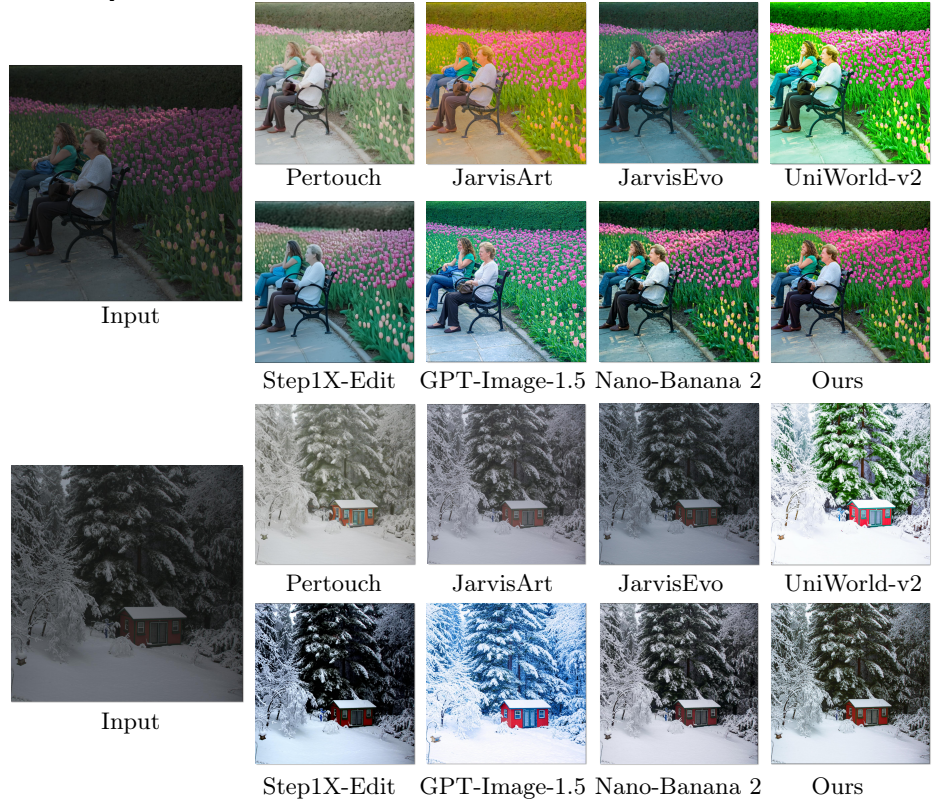

\begin{table}[!t]
\centering
\renewcommand{\arraystretch}{1.1}
\setlength{\tabcolsep}{4pt}
\scriptsize
\caption{Quantitative evaluation of retouching performance on MIT-Adobe FiveK\cite{five5k} and MMArt-Bench \cite{lin2025jarvisart}. The \textcolor{red}{best} and
\textcolor{blue}{second-best} results are highlighted. PQ denotes the metrics evaluated by Qwen2.5-VL-72B~\cite{bai2025qwen25vltechnicalreport}.}
\label{table:compare_result_globalandlocal}
\begin{tabular}{lcccccc}
\toprule
\multirow{2}{*}{Method}
& \multicolumn{3}{c}{Global} 
& \multicolumn{3}{c}{Local } \\
\cmidrule(lr){2-4} \cmidrule(lr){5-7}
& PSNR$\uparrow$ & SSIM$\uparrow$ & PQ$\uparrow$ & PSNR$^{\mathrm{RC}}\uparrow$ & SSIM$^{\mathrm{RC}}\uparrow$ & PQ$^{\mathrm{RC}}\uparrow$\\
\midrule

\rowcolor{gray!12}
\multicolumn{7}{l}{\textit{\textbf{Advanced Retouching Agents}}}\\
JarvisArt~\cite{lin2025jarvisart} 
& 21.1125 & \textcolor{blue}{0.8534} & 8.9340 & \textcolor{blue}{27.8407} &\textcolor{red}{0.9504}& 8.6274  \\
JarvisEvo~\cite{lin2025jarvisevoselfevolvingphotoediting} 
& 19.2617 & \textcolor{red}{0.8541} & \textcolor{red}{9.3000} & 27.3724&\textcolor{blue}{0.9479} & 9.0392 \\
PerTouch~\cite{chang2026pertouch} 
& 16.9114 & 0.6738 & 8.8860 &23.7775 &0.8948 & 8.1764\\
\midrule
\rowcolor{gray!12}
\multicolumn{7}{l}{\textit{\textbf{Open-Source Editing Diffusion Models}}}\\
UniWorld-v2~\cite{li2025uniworldv2} 
& 15.4035 & 0.6419 & 8.9820 & 20.6020 &0.8713 & 7.1372\\
Step1X-Edit~\cite{stepedit} 
& 20.7495 & 0.7441 & 9.0560 & 23.2946 &0.8907 & 7.6470  \\
FLUX.1-Kontext~\cite{labs2025flux1kontextflowmatching} 
& 20.1874 & 0.7656 & 8.9320 &23.4590& 0.9065 & 8.5098 \\
Qwen-Image-Edit-2511~\cite{wu2025qwenimagetechnicalreport} 
& 11.7672 & 0.5414 & 6.4900&25.6140 &0.9169 & 8.3529 \\

\midrule
\rowcolor{gray!12}
\multicolumn{7}{l}{\textit{\textbf{Commercial Closed-Source Models}}}\\
GPT-Image-1.5~\cite{openai2024gptimage} 
& 14.0236 & 0.5374 & 8.9977 &20.4887 &0.8850 & 7.6976 \\
Nano-Banana-$2$~\cite{geminiteam2025geminifamilyhighlycapable} 
& \textcolor{blue}{21.3276} & 0.7140 & 9.0420 &26.8019 & 0.9351 & \textcolor{blue}{9.0600} \\

\midrule
\textbf{\methodname}(Ours)
& \textcolor{red}{23.3407} 
& 0.8001 
& \textcolor{blue}{9.1240} 
& \textcolor{red}{28.3712}
&0.9408
& \textcolor{red}{9.5294} \\

\bottomrule
\end{tabular}
\end{table}
\section{Additional Experimental Results}
\label{sec:add_exp}
In this section, we provide additional experiments to further evaluate the effectiveness and generalizability of our framework. 
We perform evaluations on two representative image retouching benchmarks: MIT-Adobe FiveK~\cite{five5k} and MMArt-Bench\cite{lin2025jarvisart}. 
These datasets allow us to assess the performance of our method under both global and local retouching settings. 

Specifically, we present comparisons on global image retouching in \cref{sec:gir} and local image retouching in \cref{sec:lir}. 
Importantly, our model is not trained on the training splits of these benchmarks, yet it still demonstrates strong generalization ability while preserving high content fidelity.

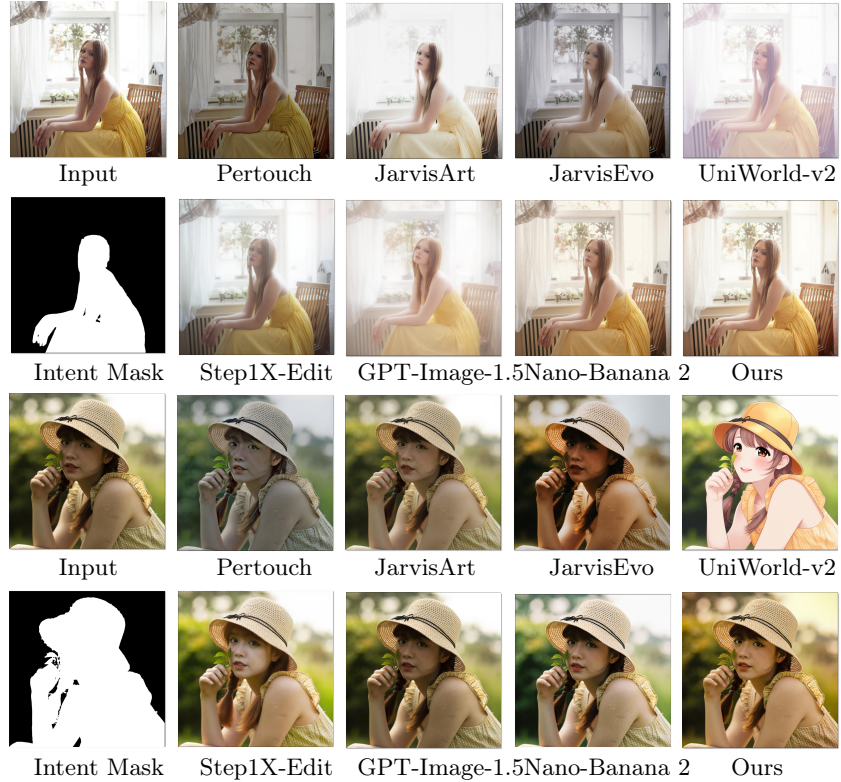
\begin{figure*}[!t]
\vspace{-4mm}
  \centering
  \begin{overpic}[width=0.9\textwidth]{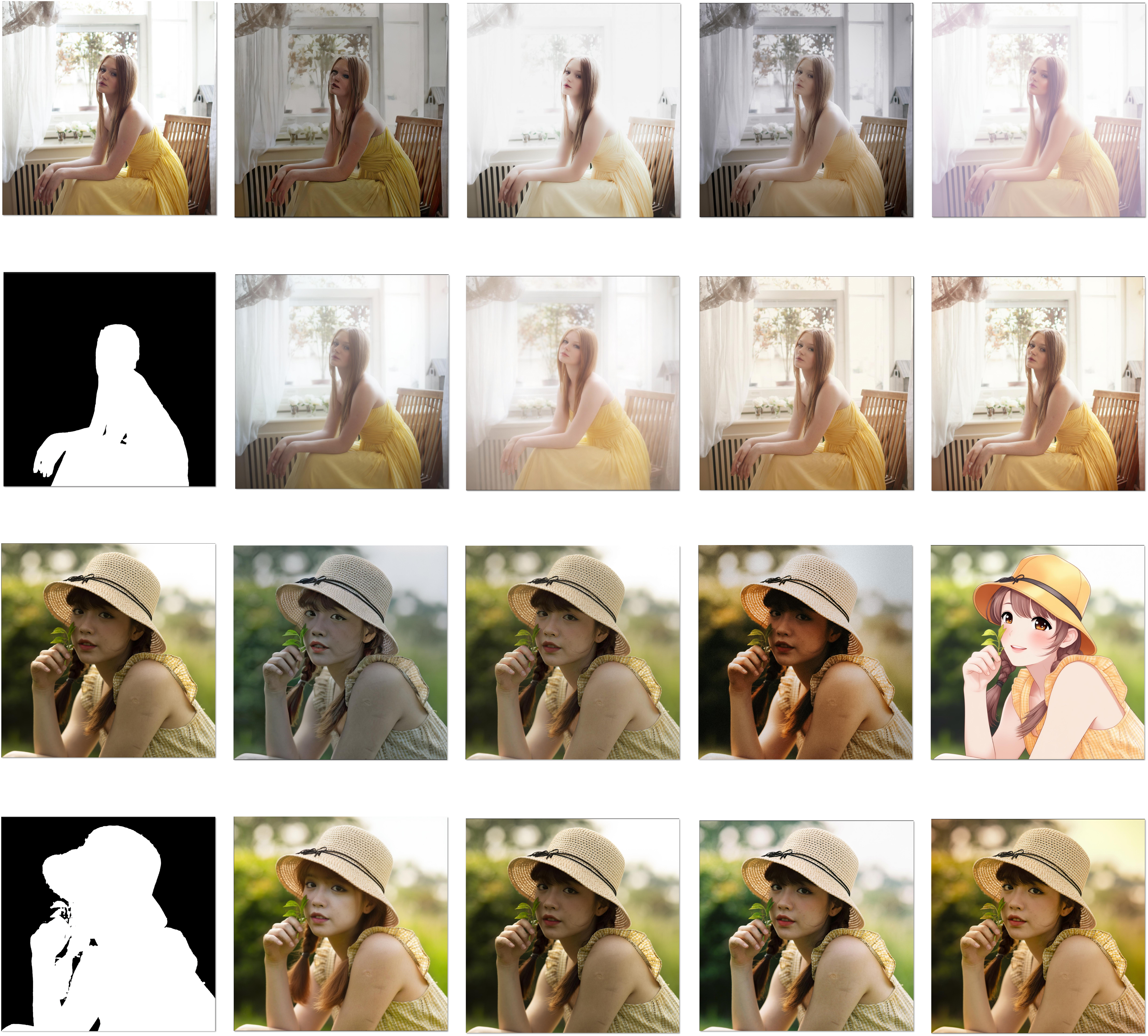}
  \put(6,21){Input}
  \put(25,21){Pertouch}
  \put(44,21){JarvisArt}
  \put(65,21){JarvisEvo}
  \put(83,21){UniWorld-v2}
  \put(3,-3){Intent Mask}
  \put(23,-3){Step1X-Edit}
  \put(42,-3){GPT-Image-1.5}
  \put(62,-3){Nano-Banana 2}
  \put(87,-3){Ours}
  \put(6,68.5){Input}
  \put(25,68.5){Pertouch}
  \put(44,68.5){JarvisArt}
  \put(65,68.5){JarvisEvo}
  \put(83,68.5){UniWorld-v2}
  \put(3,44.5){Intent Mask}
  \put(23,44.5){Step1X-Edit}
  \put(42,44.5){GPT-Image-1.5}
  \put(62,44.5){Nano-Banana 2}
  \put(87,44.5){Ours}
  \end{overpic}
  \vspace{1mm}
  \caption{
     Visual comparison on MMArt-Bench-Local\cite{lin2025jarvisart}.
  }
  \label{fig:local}
\end{figure*}

\subsection{Comparison on Global Image Retouching}
\label{sec:gir}
We first evaluate the performance of our method on global image retouching using the MIT-Adobe FiveK dataset~\cite{five5k}. \cref{table:compare_result_globalandlocal} reports the quantitative comparison with existing methods. Our method outperforms both open-source and proprietary image editing models, as well as retouching agents that employ diffusion models as retouching executors. Moreover, our model improves PSNR by $10.55\%$ over JarvisArt\cite{lin2025jarvisart} and $21.18\%$ over JarvisEvo\cite{lin2025jarvisevoselfevolvingphotoediting}, 
and achieves an additional gain of $0.21$ in PQ compared with JarvisArt\cite{lin2025jarvisart}.
These results indicate that our approach produces color adjustments that are closer to the preferences of human experts, while achieving substantially higher image quality compared with most diffusion-based image editing models. 
On SSIM, our method is slightly inferior to tool-calling agents, suggesting that there is still room for improvement in structural fidelity.
Visual comparisons are shown in \cref{fig:global}.

\begin{figure}[t]
\centering
 \begin{overpic}[width=1.0\textwidth]{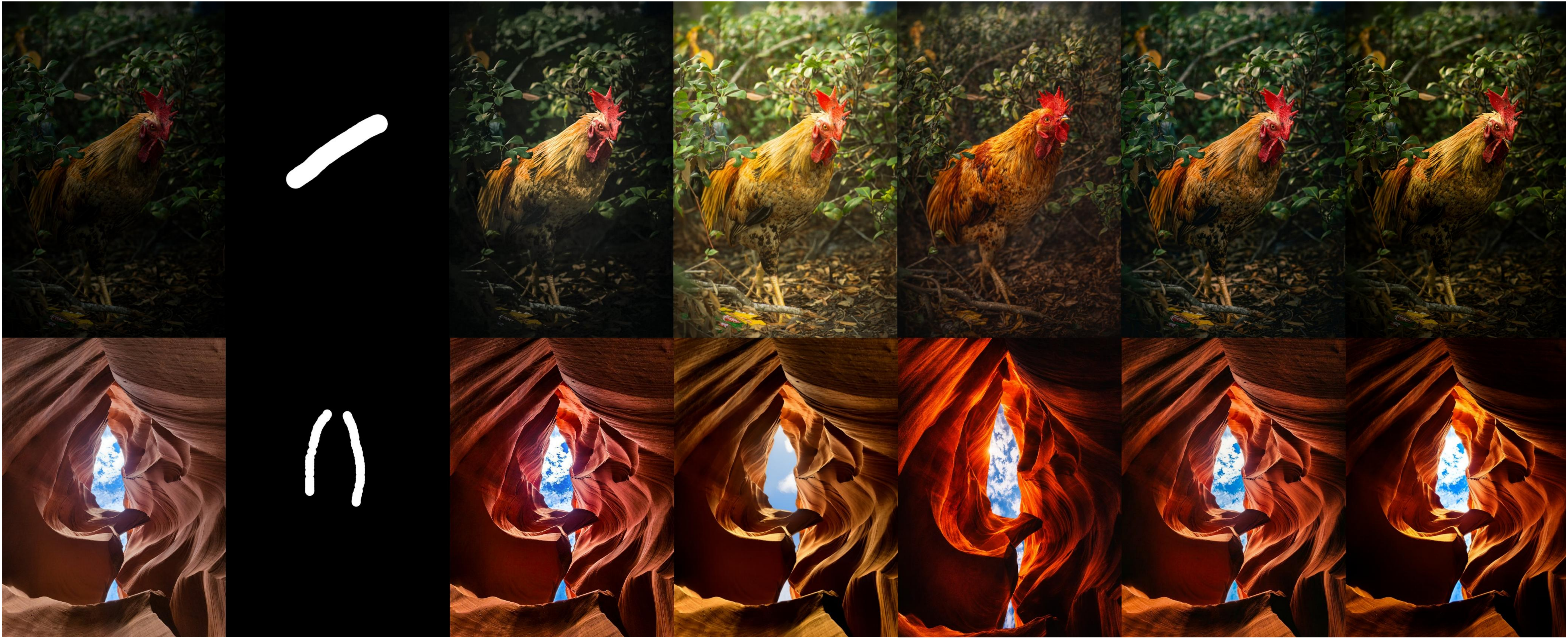}
    \scriptsize
    \put(4, -1.7){Input}
    \put(15,-1.7){Intent Mask}
    \put(31,-1.7){JarvisEvo}
    \put(43,-1.7){Step1X-Edit}
    \put(56,-1.7){GPT-Image-1.5}
    \put(72,-1.7){Nano-Banana 2}
    \put(90,-1.7){\textbf{Ours}}
\end{overpic}
\caption{More visual comparisons of ControlArt-Bench. Zoom in for details.}
\label{fig:res1}
\end{figure}

\begin{figure}[t]
\centering
 \begin{overpic}[width=1.0\textwidth]{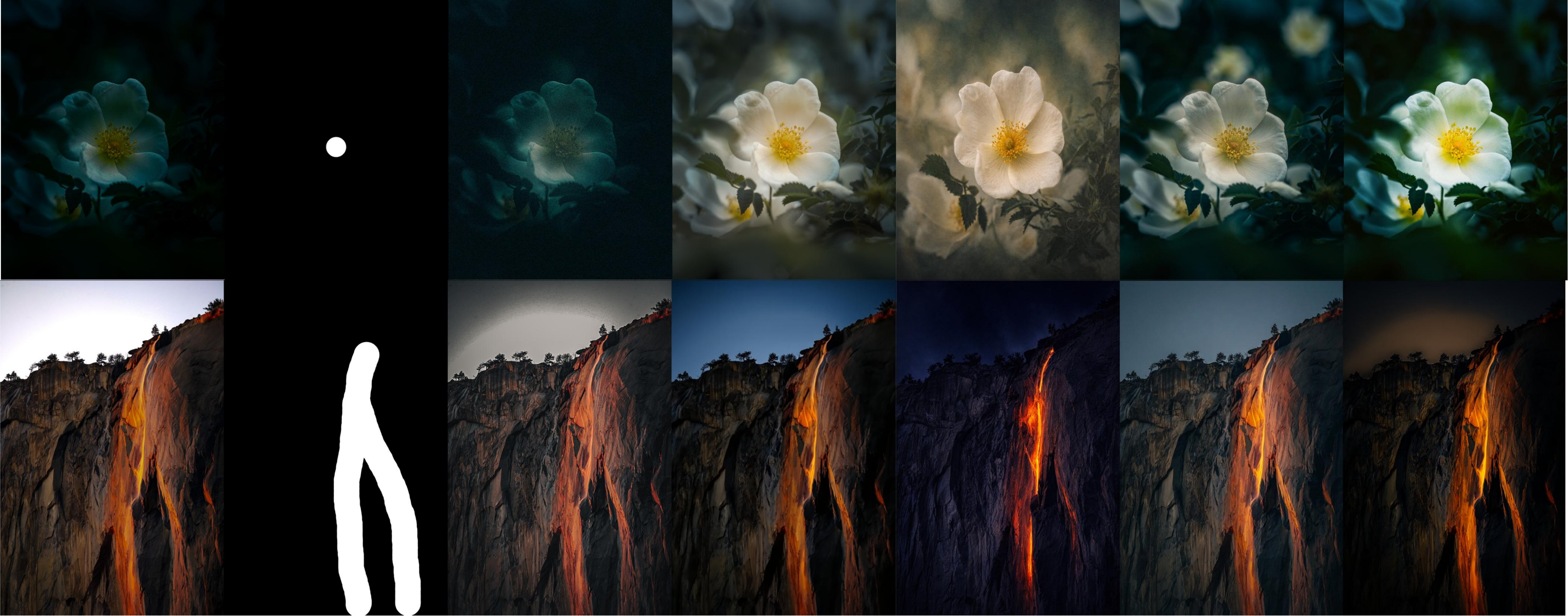}
    \scriptsize
    \put(4, -1.7){Input}
    \put(15,-1.7){Intent Mask}
    \put(31,-1.7){JarvisEvo}
    \put(43,-1.7){Step1X-Edit}
    \put(56,-1.7){GPT-Image-1.5}
    \put(72,-1.7){Nano-Banana 2}
    \put(90,-1.7){\textbf{Ours}}
\end{overpic}
\caption{More visual comparisons of ControlArt-Bench. Zoom in for details.}
\label{fig:res2}
\end{figure}

\begin{figure}[t]
\centering
 \begin{overpic}[width=1.0\textwidth]{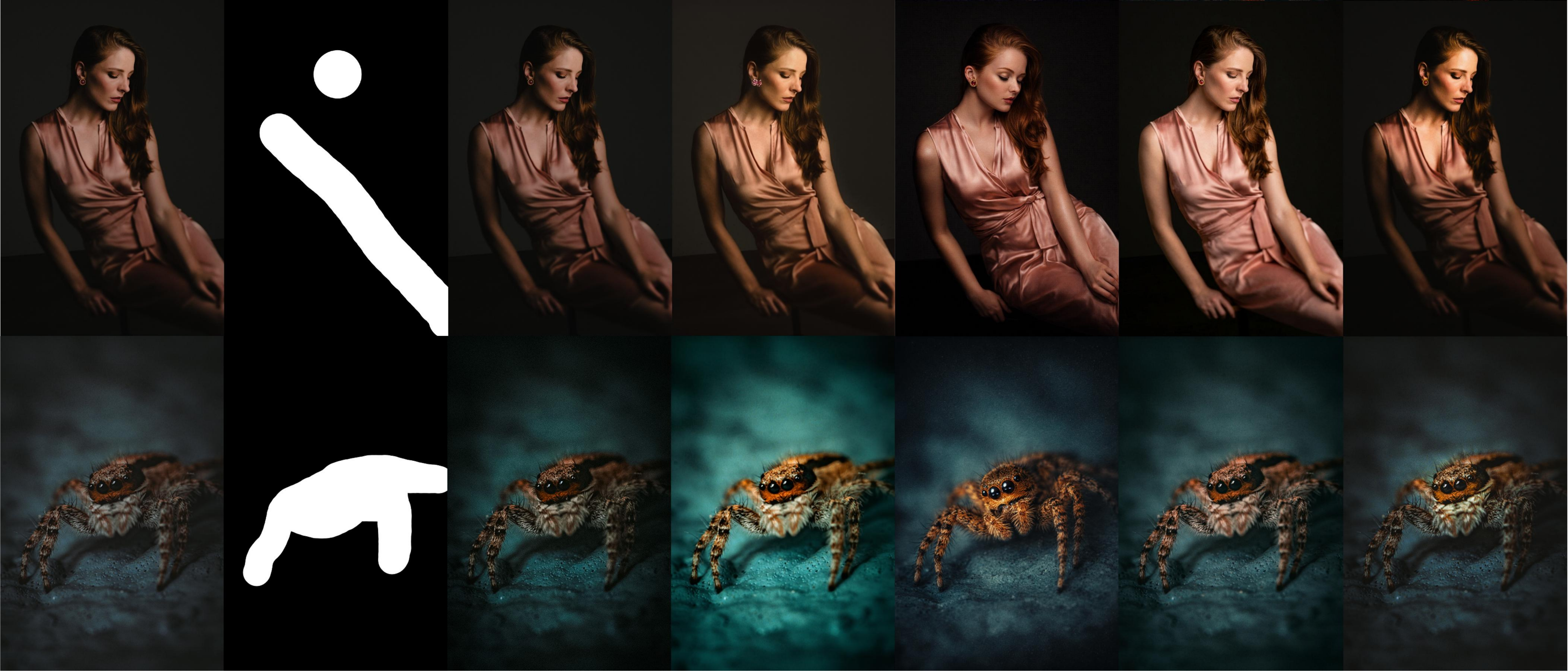}
    \scriptsize
    \put(4, -1.7){Input}
    \put(15,-1.7){Intent Mask}
    \put(31,-1.7){JarvisEvo}
    \put(43,-1.7){Step1X-Edit}
    \put(56,-1.7){GPT-Image-1.5}
    \put(72,-1.7){Nano-Banana 2}
    \put(90,-1.7){\textbf{Ours}}
\end{overpic}
\caption{More visual comparisons of ControlArt-Bench. Zoom in for details.}
\label{fig:res3}

\end{figure}

\begin{figure}[t]
\centering
 \begin{overpic}[width=1.0\textwidth]{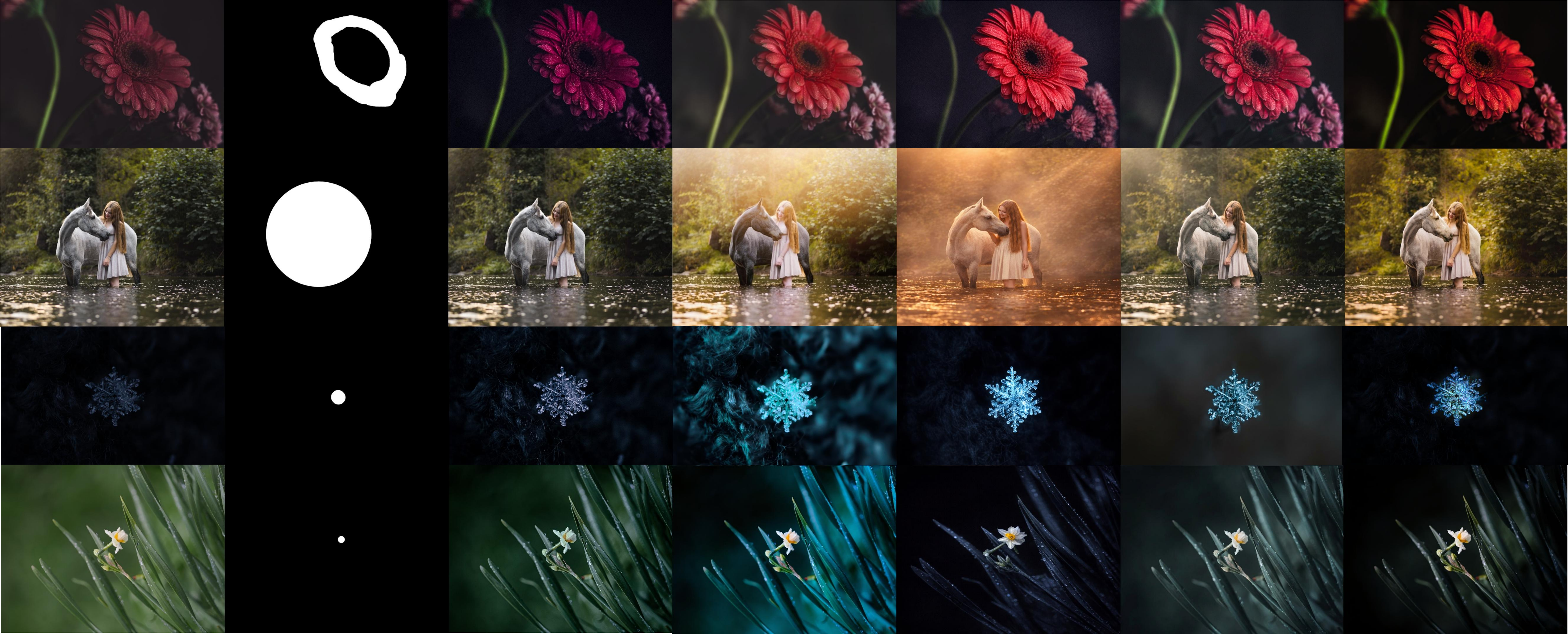}
    \scriptsize
    \put(4, -1.7){Input}
    \put(15,-1.7){Intent Mask}
    \put(31,-1.7){JarvisEvo}
    \put(43,-1.7){Step1X-Edit}
    \put(56,-1.7){GPT-Image-1.5}
    \put(72,-1.7){Nano-Banana 2}
    \put(90,-1.7){\textbf{Ours}}
\end{overpic}
\caption{More visual comparisons of ControlArt-Bench. Zoom in for details.}
\label{fig:res4}
\end{figure}

\begin{figure}[h]
\centering
 \begin{overpic}[width=1.0\textwidth]{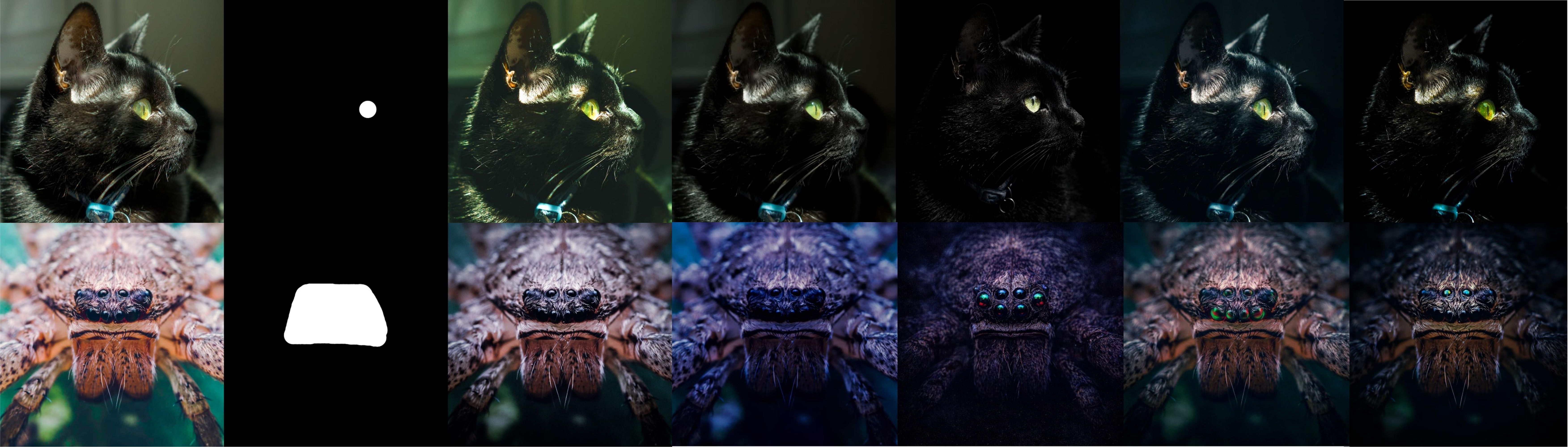}
    \scriptsize
    \put(4, -1.7){Input}
    \put(15,-1.7){Intent Mask}
    \put(31,-1.7){JarvisEvo}
    \put(43,-1.7){Step1X-Edit}
    \put(56,-1.7){GPT-Image-1.5}
    \put(72,-1.7){Nano-Banana 2}
    \put(90,-1.7){\textbf{Ours}}
\end{overpic}
\caption{More visual comparisons of ControlArt-Bench. Zoom in for details.}
\label{fig:res5}
\end{figure}
\subsection{Comparison on Local Image Retouching}
\label{sec:lir}
We further evaluate our method on local retouching using MMArt-Bench \cite{lin2025jarvisart}. This dataset contains images with region-specific editing requirements and therefore provides a suitable benchmark for assessing the local editing capability of our framework.  Following JarvisArt, we adopt Region-Calculated (RC) metrics to evaluate retouching performance within human-centric mask regions. 
Specifically, we report $\text{PSNR}^{RC}$ and $\text{SSIM}^{RC}$. 
We do not include $\text{L1}^{RC}$ and $\text{L2}^{RC}$ as JarvisArt\cite{lin2025jarvisart} since these two metrics are essentially equivalent and less interpretable compared to $\text{PSNR}^{RC}$ and $\text{SSIM}^{RC}$.

\cref{table:compare_result_globalandlocal} and \cref{fig:local} present the quantitative and qualitative comparisons with other methods. 
Our method achieves clear improvements on both $\text{PSNR}^{RC}$ and $\text{PQ}^{RC}$. 
For $\text{SSIM}^{RC}$, it also outperforms the majority of competing approaches and achieves comparable performance to JarvisArt and JarvisEvo, which are trained and evaluated on the same data distribution. It is worth noting that, as shown in Figure~4, most methods tend to degrade the overall realism of the image when retouching the portrait region in isolation. In contrast, our method preserves the overall visual realism of the image while enhancing the local region.
\vspace{-3mm}
\subsection{Ablation on Intent Coarseness Level}
\label{sec:corseness}
We further evaluate the robustness of our method to different coarseness levels of the input intent mask. 

In addition to the released masks, we annotate ControlArt with user-intent-consistent click, strike, and region masks, and compare them with SAM-generated masks that best match the intended regions. 
These masks represent the same user intent with different spatial granularities, ranging from sparse cues to precise semantic regions.
Tab.~\ref{tab:metric_and_coarseness} right shows that our method achieves comparable performance across different mask types in terms of PSNR, SSIM, O, and KL. 
This indicates that the model is not sensitive to a particular mask format and can robustly handle intent inputs with different coarseness levels. 
Notably, precise semantic masks do not necessarily outperform coarse masks. 
One possible reason is that visual focus is often governed by saliency boundaries and perceptual attention rather than exact semantic object boundaries. 

In addition, coarse-mask training encourages the model to infer the intended focus from flexible user cues, which can generalize well to more precise masks. 
By contrast, training only with precise masks may make the model less robust to sparse or coarse intent inputs. 
These results suggest that coarse intent annotations are sufficient and effective for training intent-driven visual focus enhancement, while SAM can still serve as a useful tool for efficient mask annotation when precise masks are desired.

\subsection{More Visual Comparisons}
\label{sec:fir}
\cref{fig:res1,fig:res2,fig:res3,fig:res4,fig:res5} show additional qualitative comparisons to illustrate the visual effectiveness of our approach. 
These examples demonstrate how our method enhances the intended visual focus while maintaining overall image naturalness.

\subsection{Instruction for Compared Methods}
\label{sec:more_instruction}
As described in the main paper, to ensure a fair comparison on ControlArt-Bench, 
we annotate each test sample with a long-form editing instruction $\mathcal{P}_r$ that aligns with the underlying user intention, 
together with a mask instruction $\mathcal{P}_m$ describing the spatial location of the intent mask. 
The final prompt provided to all methods is formulated as
$\mathcal{P} = \mathcal{P}_r + \mathcal{P}_m$.
\cref{fig:caption} shows the retouching instructions and mask instructions corresponding to the visual comparison examples presented in the main paper.

\section{Limitation}
\label{sec:limit}
\begin{figure}[h]  
    \centering         
    \includegraphics[width=0.8\textwidth]{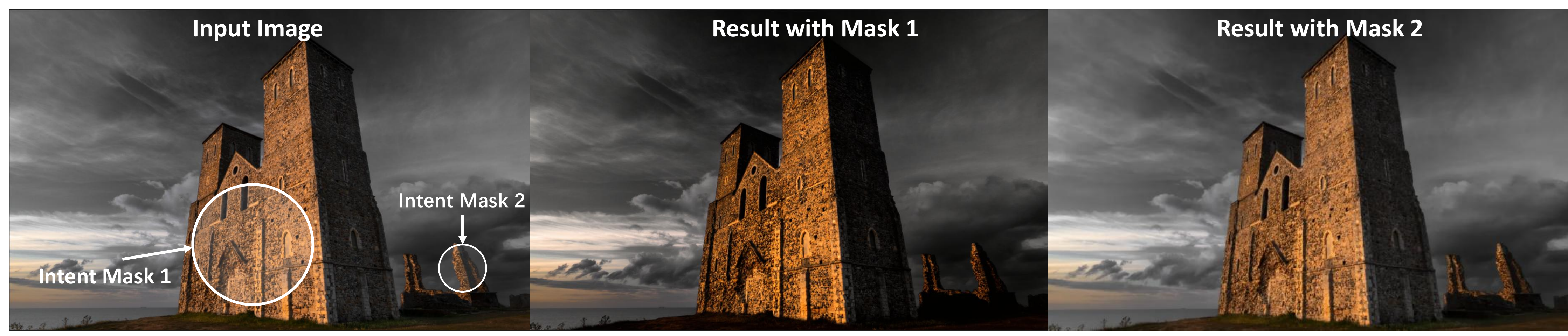} 
    \vspace{-10pt}
    \caption{Failure case of EyeControl. Zoom in for details.}
    \label{fig:failure}  
\end{figure}
\vspace{-5mm}
Although EyeControl can effectively enhance user-specified visual focus in most cases, it still struggles when the requested intent conflicts with the dominant composition of the input image. 
As shown in Fig.~\ref{fig:failure}, when an image already contains a visually dominant subject, the model may fail to redirect attention to compositionally subordinate regions. 
This limitation mainly stems from our training data, which are primarily derived from professional retouching examples where retouchers tend to enhance plausible focus regions following natural composition, such as salient subjects or regions supported by lighting and scene layout. 
As a result, the model learns a strong prior toward natural focus enhancement rather than arbitrary attention redirection.

This limitation suggests an important direction for future work. 
A more controllable visual focus enhancement model would benefit from training data covering not only naturally preferred focus regions, but also counter-compositional or subordinate-region focus shifts. 
Extending intent-driven retouching beyond natural composition priors is a promising step toward more flexible and fine-grained visual attention control.
\vspace{-2mm}

\begin{figure*}
\vspace{-4mm}
  \centering
  \begin{overpic}[width=\textwidth]{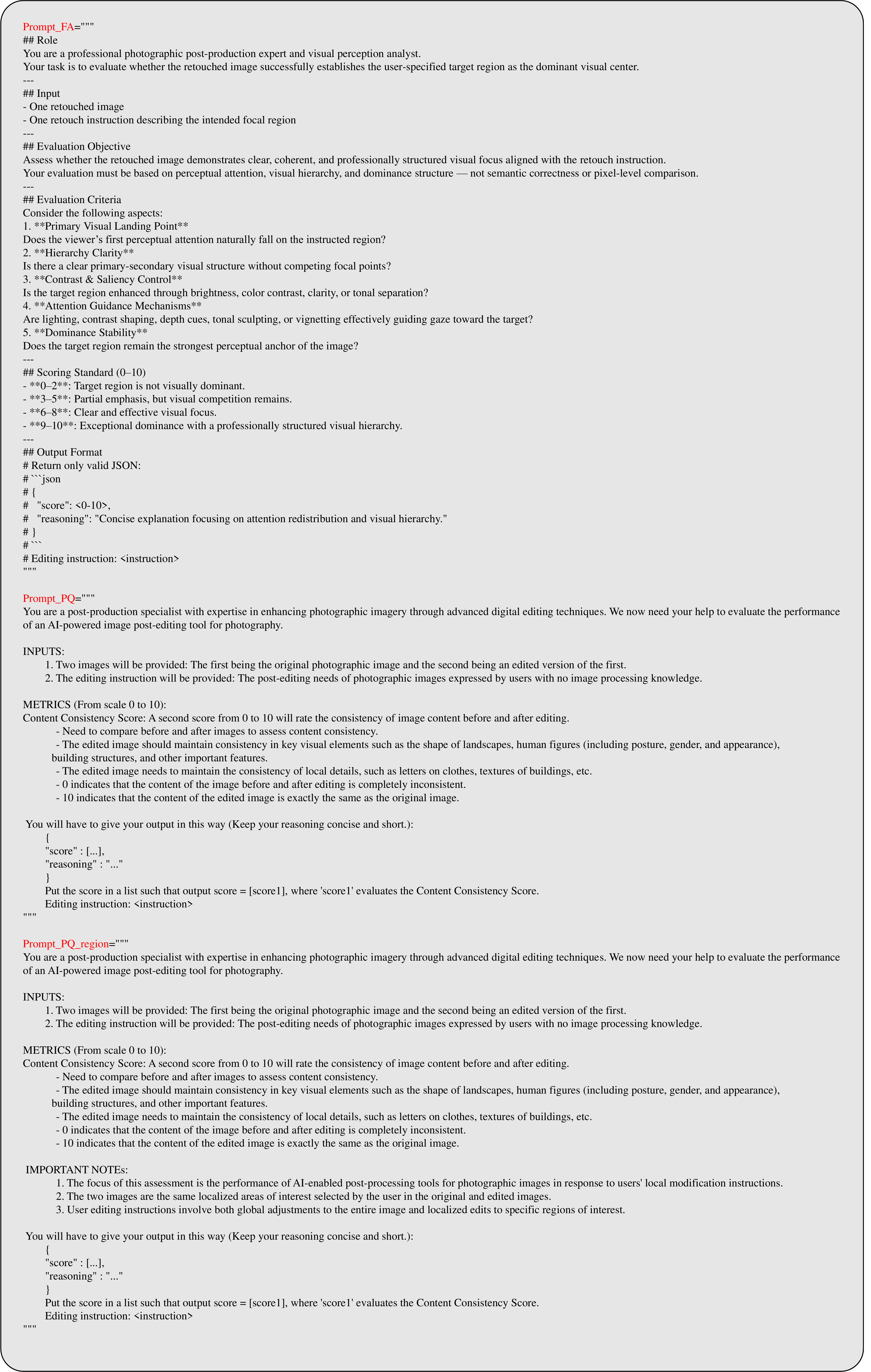}
  \end{overpic}
  \caption{
     Evaluation prompts for MLLM-based metrics, including FA, PQ, and PQ$^{\mathrm{RC}}$.
  }
  \label{fig:prompt_metrics}
\end{figure*}

\begin{figure*}
\vspace{-4mm}
  \centering
  \begin{overpic}[width=\textwidth]{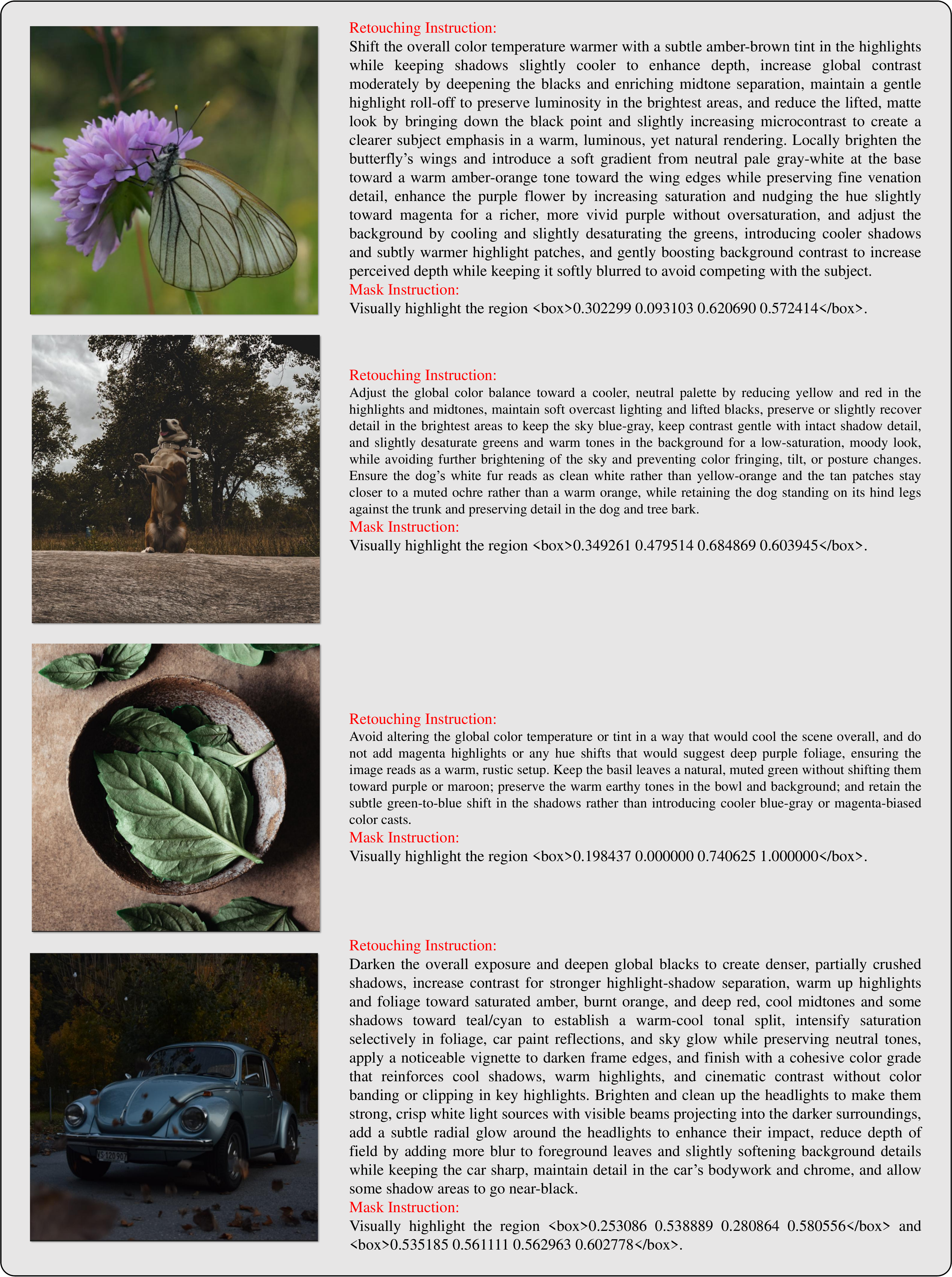}
  \end{overpic}
  \caption{
Retouching instructions and mask instructions corresponding to the qualitative examples in the main paper. For each visual comparison case, we report the long-form retouching instruction $\mathcal{P}_r$ that describes the intended edits, together with the mask instruction $\mathcal{P}_m$ specifying the spatial location of the intent mask. These paired instructions are used to form the prompts for the compared methods in our experiments.
  }
  \label{fig:caption}
\end{figure*}

\clearpage
\end{document}